\documentclass[fleqn]{article}

\usepackage{microtype}
\usepackage{graphicx}
\usepackage{booktabs} 
\usepackage[utf8]{inputenc}

\usepackage{amsmath,amsfonts,bm}

\def\figref#1{figure~\ref{#1}}

\def\secref#1{section~\ref{#1}}

\def\eqref#1{equation~\ref{#1}}

\def\1{\bm{1}}

\DeclareMathAlphabet{\mathsfit}{\encodingdefault}{\sfdefault}{m}{sl}
\SetMathAlphabet{\mathsfit}{bold}{\encodingdefault}{\sfdefault}{bx}{n}

\newcommand{\R}{\mathbb{R}}

\newcommand{\KL}{D_{\mathrm{KL}}}

\newcommand{\ubar}[1]{\underaccent{\bar}{#1}}

\newtheorem{theorem}{Theorem}

\newcommand\norm[1]{\left\lVert#1\right\rVert}

\newcommand{\s}[1]{\tilde{#1}}
\newcommand{\hcm}{\hspace{0.5cm}}
\newcommand{\ocm}{\hspace{1cm}}
\newcommand{\F}[1]{{\cal{F}}\cb{#1}}
\newcommand{\ave}[2]{{\mathbb{E}_{#2}\sq{#1}}}
\newcommand{\david}[1]{({\color{red}{#1 - David}})}

\newcommand{\peter}[1]{({\color{blue}{#1 - Peter}})}

\newcommand{\abs}[1]{\left| {#1} \right|}
\newcommand{\cb}[1]{\left\{ {#1} \right\}}
\newcommand{\br}[1]{\left( {#1} \right)}
\newcommand{\sq}[1]{\left[ {#1} \right]}

\renewcommand{\KL}[2]{\textrm{KL}\!\br{#1\vert\vert#2}}

\newcommand{\conv}[2]{\br{#1 * #2}}
\renewcommand{\div}[2]{\textrm{D}\!\br{#1\vert\vert#2}}
\newcommand{\fdiv}[2]{\textrm{D}_f\!\br{#1\vert\vert#2}}
\newcommand{\sdiv}[2]{\tilde{\textrm{D}}\!\br{#1\vert\vert#2}}
\newcommand{\sfdiv}[2]{\tilde{\textrm{D}}_f\!\br{#1\vert\vert#2}}
\newcommand{\sKL}[2]{\widetilde{\textrm{KL}}\!\br{#1\vert\vert#2}}
\newcommand{\const}{{const.}}
\newcommand{\trace}[1]{\text{Tr}\br{#1}}

\newcommand{\supp}[1]{{\text{supp}}\br{#1}}
\renewcommand{\figref}[1]{figure(\ref{#1})}
\renewcommand{\secref}[1]{section(\ref{#1})}
\newcommand{\appref}[1]{appendix(\ref{#1})}

\newcommand{\beq}{\begin{equation}}
\newcommand{\eeq}{\end{equation}}

\newcommand{\ndist}[3]{{\cal{N}}\br{#1\thinspace\vline\thinspace #2,#3}}
\newcommand{\trans}{^{\textsf{T}}}
\newcommand{\ntrans}{^{-\textsf{T}}}
\newcommand{\sett}[1]{\mathcal{\uppercase{#1}}}

\newcommand{\qcm}{\hspace{0.2cm}}

\usepackage{amsmath, amssymb}
\usepackage{hyperref,tikz}
\usepackage{url}
\usepackage{subfig}
\usepackage{url}
\usepackage{floatrow}
\usepackage[export]{adjustbox}
\usepackage{wrapfig}
\usepackage{sidecap}
\usepackage{nccmath}
\usepackage{hyperref}
\usepackage{url}
\usepackage[utf8]{inputenc} 
\usepackage[T1]{fontenc}    
\usepackage{hyperref}       
\usepackage{url}            
\usepackage{amsfonts}       
\usepackage{nicefrac}       
\usepackage{amsthm}
\usepackage{accents}
\newtheorem{thm}{Theorem}
\newtheorem{lemma}[theorem]{Lemma}
\theoremstyle{definition}

\usepackage{hyperref}

\renewcommand{\eqref}[1]{eq(\ref{#1})}
\newcommand{\sz}[1]{\text{dim}(#1)}

\usepackage[accepted]{icml2020}

\icmltitlerunning{Spread Divergence}
\begin{document}

\twocolumn[
\icmltitle{Spread Divergence}



\begin{icmlauthorlist}

\icmlauthor{Mingtian Zhang}{ucl}
\icmlauthor{Peter Hayes}{ucl}
\icmlauthor{Tom Bird}{ucl}
\icmlauthor{Raza Habib}{ucl}
\icmlauthor{David Barber}{ucl}
\end{icmlauthorlist}


\icmlaffiliation{ucl}{Department of Computer Science, University College London, UK}

\icmlcorrespondingauthor{Mingtian Zhang}{mingtian.zhang.17@ucl.ac.uk}

\icmlkeywords{Machine Learning, ICML}

\vskip 0.3in
]



\printAffiliationsAndNotice{} 

\begin{abstract}
For distributions $\mathbb{P}$ and $\mathbb{Q}$ with different supports or undefined densities, the divergence $\textrm{D}(\mathbb{P}||\mathbb{Q})$ may not exist. We define a Spread Divergence $\tilde{\textrm{D}}(\mathbb{P}||\mathbb{Q})$  on modified $\mathbb{P}$ and $\mathbb{Q}$ and describe sufficient conditions for the existence of such a divergence. We demonstrate how to maximize the discriminatory power of a given divergence by parameterizing and learning the spread. We also give examples of using a Spread Divergence to train implicit generative models, including linear models (Independent Components Analysis) and non-linear models (Deep Generative Networks). 
\end{abstract}

\section{Introduction}
For distributions $\mathbb{P}$ and $\mathbb{Q}$, a divergence $\div{\mathbb{P}}{\mathbb{Q}}$ is a measure of their difference \citep{Dragomir2005} provided it satisfies the properties\footnote{Two distributions being equal $\mathbb{P}=\mathbb{Q}$ can be interpreted to mean that the cdfs (cumulative distribution functions) of the two distributions match.}
\beq
\div{\mathbb{P}}{\mathbb{Q}}\geq 0 \qcm\text{and} \qcm \div{\mathbb{P}}{\mathbb{Q}} = 0 \qcm \Leftrightarrow \qcm \mathbb{P}=\mathbb{Q}.
\eeq
For absolutely continuous  distributions\footnote{A distribution is absolutely continuous if its cdf is absolutely continuous. In this case, it has a density function. See also \citet[p.~172]{tao2011introduction}.} $\mathbb{P}$, $\mathbb{Q}$ and probability density functions $p(x)$, $q(x)$,  the $f$-divergence is 
\beq
\fdiv{\mathbb{P}}{\mathbb{Q}}\equiv\fdiv{p}{q} = \ave{f\br{\frac{p(x)}{q(x)}}}{q(x)},
\eeq
where $f(x)$ is a convex function with $f(1)=0$. When $p$ and $q$ have the same support\footnote{We use $a.e.$ to represent `almost everywhere', see  \citet[p.~18]{durrett2019probability} for a definition.},
\beq 
\fdiv{p}{q} =0\Leftrightarrow p=q \; a.e. \Leftrightarrow\mathbb{P}=\mathbb{Q},
\eeq
see \citep{fdiv1,fdiv2}. A well-known $f$-divergence is the Kullback-Leibler (KL) divergence
\beq
\KL{p}{q} = \ave{\log{\frac{p(x)}{q(x)}}}{p(x)}.
\label{eq:kl:def}
\eeq
The KL divergence plays a key role in fitting a model $q(x)$ to data. For training data $x_1,\ldots, x_N$ generated iid from a distribution $p(x)$, a sample based approximation of \eqref{eq:kl:def} is
\beq
\KL{p}{q} = -\frac{1}{N}\sum_{n=1}^N \log q(x_n)+ \const
\label{eq:kl:approx}
\eeq
where $\sett{L}(\theta)\equiv \sum_{n=1}^N \log q(x_n)$ is the data log likelihood and $\theta$ are the parameters of the model $q(x)$.  
Hence, fitting a model $q$ to data using maximum likelihood can be viewed as minimizing the KL divergence between the model and the data generating process.

Using an $f$-divergence $\fdiv{\mathbb{P}}{\mathbb{Q}}$ for model training therefore requires (i) $\mathbb{P}$ and $\mathbb{Q}$ to have valid probability densities $p, q$ and (ii) $p$ and $q$ to have common support. However, these requirements are not satisfied in some important machine learning applications, in particular in implicit generative models, as we discuss below.

%
%

\subsection{Implicit Generative Models}

\begin{figure}[t]
\includegraphics[width=0.4\textwidth]{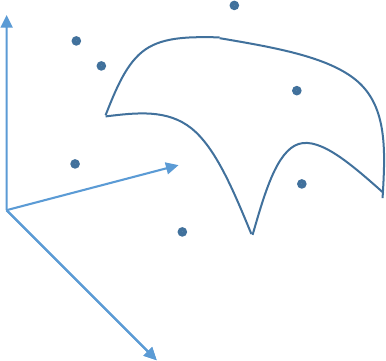}
\caption{The figure shows an implicit generative model with data $\sz{X}=3$ and latent $\sz{Z}=2$. The model can only generate data (dots) on a lower dimensional manifold in $X$ space. The likelihood of generating a data point off the manifold is zero meaning that the likelihood will in general be a non-continuous function of the parameters defining the manifold.
\label{fig:intro_implict_model}}
\end{figure}

Implicit generative models are of considerable recent interest in machine learning. These take the form of a latent variable model $q(x) = \int q(x|z)q(z)dz$, but with a deterministic output distribution $q(x|z)=\delta(x-g_\theta(z))$. We can define the generalised density\footnote{We use `generalised density' to include implicit generative models. This includes also the special case $p(x)=\delta(x-\mu)$, where $p(x)$ represents a (Dirac) delta distribution with support $\{\mu\}$. In general we cannot define an $f$-divergence based on generalised densities, but can do so if the distribution is absolutely continuous - see \secref{sec:general:stationary} for details.}, for the model $\mathbb{Q}$ as
\beq
q(x) = \int \delta\br{x-g_\theta(z)}q(z)dz, 
\label{eq:implicit}
\eeq
where $z$ represents the value of latent variable $Z$, $x$ the value of observation variable $X$ and $\delta(x)$ is the Dirac delta function. In the common setting where the latent dimension is lower than the observation dimension $\sz{Z}<\sz{X}$, the model $q(x)$ only has limited support, see \figref{fig:intro_implict_model}. Strictly speaking, this does not define a density over the real space $\mathbb{R}^{\sz{X}}$ since $X$ is not absolutely continuous. Nevertheless, this does define a distribution $\mathbb{Q}$.  To generate a sample from $\mathbb{Q}$, one first samples $z$ from $q(z)$ and then passes this through the `generator' function $g_\theta(z)$. 

In this case, gradient based maximum likelihood learning is problematic since $\sett{L}(\theta)$ is typically not a continuous function of $\theta$\footnote{For a point $x_n$ that is not on the model manifold, $q(x_n)=0$. As we adjust $\theta$ such that $x_n$ becomes on the manifold, $q(x_n)$ will typically increase to a finite non-zero value, meaning that the (log) likelihood is discontinuous in $\theta$.}. Furthermore, the Expectation Maximisation algorithm \citep{dempster1977maximum} is  not available for models of the form in \eqref{eq:implicit} since EM assumes that $\log q(x|z)$ is well defined, which is not the case when $q(x|z)=\delta\br{x-g_\theta(z)}$ \citep{Bermond99approximatelikelihood}. Equally, in \eqref{eq:kl:def} the ratio $p(x)/q(x)$ may represent a division by zero; the KL divergence between the model and the data generating process is thus ill-defined. 

It is natural to consider transforming two distributions $\mathbb{P}$ and $\mathbb{Q}$ to have common support by using a mixture  model $\tilde{\mathbb{P}}=\alpha\mathbb{P}+(1-\alpha)\mathbb{N}$, $\tilde{\mathbb{Q}}=\alpha\mathbb{Q}+(1-\alpha)\mathbb{N}$, where $\mathbb{N}$ is an absolutely continuous noise distribution with full support. However, as we explain in \appref{app:mixture:divergence} this approach is not useful in the context of training implicit generative models since $\tilde{\mathbb{Q}}$ does not have a density which can be numerically evaluated.

\subsubsection{Model noise is not enough\label{sec:model:not:enough}}
A common approach to enable maximum likelihood to be used to train implicit generative models is to simply add noise to the model so that it has full support (and a valid density), see for example \citep{wu2016}.  However, this approach does not guarantee a consistent estimator. To see this, consider the simple implicit generative model $\mathbb{Q}$ with generalised density
\beq
q(x) = \int \delta\br{x-z\theta_q}\ndist{z}{0}{1}dz,
\eeq
where the latent $Z$ is univariate and $\sz{X}>1$. Here the vector $\theta_q$ defines a one-dimensional line in the $X$-space. For $D$-dimensional $X$, adding independent Gaussian noise with mean zero and isotropic covariance $\sigma^2 I_D$ to $X$ results in the noised distribution with density
\beq
\tilde{q}(x) = \ndist{x}{0_D}{\Sigma}, \hcm \Sigma \equiv \theta_q\theta_q\trans + \sigma^2 I_D.
\eeq
For observed training data $x_1,\ldots, x_N$ the log likelihood under this model is
\beq
\frac{1}{N}\sett{L}(\theta_q) = -\frac{1}{N}\sum_{n=1}^N x_n\trans \Sigma^{-1}x_n - \log\det{\Sigma} + \const
\label{eq:scaled:log:lik}
\eeq
We assume that the training data $x_n$ is iid sampled from the distribution $\mathbb{P}$ with generalised 
density 
\beq
p(x) = \int \delta\br{x-z\theta_p}\ndist{z}{0}{1}dz.
\eeq
Hence $\mathbb{P}$ and $\mathbb{Q}$ are from the same parametric distribution but with different parameters. By the law of large numbers, in the large $N$ limit, the log likelihood \eqref{eq:scaled:log:lik} tends to
\beq
-\theta_p\trans\Sigma^{-1}\theta_p - \log\det\Sigma + \const
\label{eq:log:lik:limit}
\eeq
which has an optimum when\footnote{$\theta_p^2$ is shorthand for the squared length $\theta_p\trans\theta_p=||\theta_p||^2_2$.} (see \appref{sec:annealing:noise})
\beq
\theta_q = \sqrt{\frac{\theta_p^2-\sigma^2}{\theta_p^2}}\theta_p.
\eeq
Thus adding noise to the model $\mathbb{Q}$ and training using maximum likelihood does not form a consistent estimator; it has an optimum at $\theta_q\neq \theta_p$, resulting in an incorrect estimate of the data generating process. In \appref{sec:annealing:noise} we explain why annealing the noise $\sigma^2$ towards zero during numerical optimisation will not directly heal this problem. 

Another well-known failure case of trying to learn an implicit generative model by adding noise only to the model is deterministic ICA, which we discuss at length in \secref{sec:ica}.

For this reason alternative (non-likelihood, non-KL) approaches to measure the difference between distributions are commonly used in training implicit generative models (see for example \citet{shakir16}), such as Maximum Mean Discrepancy \citep{gretton2012kernel} and Wasserstein distance \citep{arjovsky2017wasserstein,peyre2019computational}. In the next section we introduce the Spread Divergence which defines a valid divergence even when the supports of the distributions do not match or the distributions do not have a valid density. As we will demonstrate, the Spread Divergence allows one to use maximum likelihood based approaches to train implicit generative models, whilst resulting in a consistent estimator. 



\section{Spread Divergence}


For distributions $\mathbb{Q}$ and $\mathbb{P}$ with generalised 
densities $q(x)$ and $p(x)$ we first need to define $\s{q}(y)$ and $\s{p}(y)$ that (i) are valid probability density functions and (ii) have the same support. 
In contrast to simply noising $\mathbb{Q}$ we define `noisy' densities for both distributions
\beq
\s{p}(y) = \int p(y{\mid}x)p(x)dx, \hspace{2mm}\s{q}(y) = \int p(y{\mid}x)q(x)dx
\label{eq:sdist:def}
\eeq
The `noise' $p(y{\mid}x)$ must `spread' $\mathbb{P}$ and $\mathbb{Q}$ such that $\s{p}(y)$ and $\s{q}(y)$ satisfy the above two reauirements. The choice of $p(y|x)$ must also ensure that $\div{\s{p}}{\s{q}} = 0 \Leftrightarrow \mathbb{P}=\mathbb{Q}$. If we can define the noise appropriately, this would allow us to define the Spread Divergence 
\beq
\sdiv{\mathbb{P}}{\mathbb{Q}} \equiv \div{\s{p}}{\s{q}},
\eeq
which satisfies the divergence requirement $\sdiv{\mathbb{P}}{\mathbb{Q}}\geq 0$ and  $\sdiv{\mathbb{P}}{\mathbb{Q}}=0 \Leftrightarrow \mathbb{P}=\mathbb{Q}$. In the following we discuss appropriate choices for the noise distribution $p(y|x)$. We focus on stationary spread noise $p(y|x)=k(y-x)$ since this is simple to implement by adding independent noise to a variable. Non-stationary spread distributions can be easily constructed using a mixture of stationary noise distributions, or through Mercer kernels -- these are left for future study. The case of discrete $X$ is discussed in \appref{sec:disc: noise:rec}.
 

\begin{figure*}[t]
\begin{minipage}{\textwidth}
\vspace{-4mm}
	\centering
		\subfloat[Two delta distributions\protect.]{\includegraphics[height=0.2\linewidth]{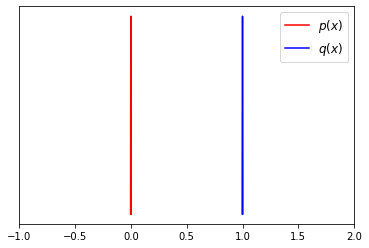}}
		\subfloat[Spreaded delta distributions.]{\includegraphics[height=0.2\linewidth]{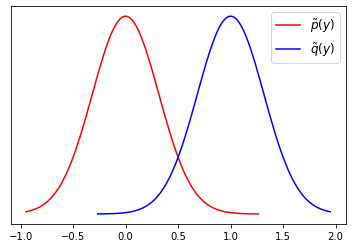}}
		\subfloat[Spread KL divergence]{\includegraphics[height=0.2\linewidth]{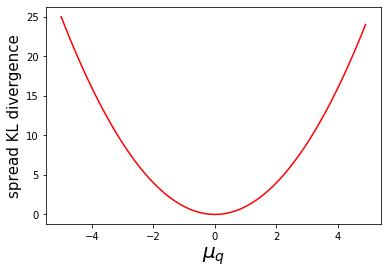}}
		\caption[]{(a) Delta distributions  $p(x)=\delta(x-\mu_p)$, $q(x)=\delta(x-\mu_q)$ where $\mu_p=0$, $\mu_q=1$. (b) Spreaded distributions $\tilde{p}(y)=\int  p(y|x)p(x)dx$, $\tilde{q}(y)=\int p(y|x)q(x)dx$, where $p(y|x)=\ndist{y}{x}{\sigma^2=0.5}$.
		(c) The spread KL divergence as a function of $\mu_q$.\label{fig:intro_toy}}
		\end{minipage} 
\end{figure*}

\subsection{Stationary Spread Divergence\label{sec:stat}}


For a random variable $X$ we define a new `stationary spread' random variable $Y$ by adding to $X$ random noise $K$. In order to ensure that $Y$ has a valid probability density function and the required support, we assume that $K$ is absolutely continuous with  density  $k(x){>}0$, $x\in\mathbb{R}$.  We then define a random variable $Y = X + K$.  In the context of \eqref{eq:sdist:def} this corresponds to using noise of the form $p(y|x)=k(y-x)$.

\subsubsection{Absolutely Continuous Distributions}
For two absolutely continuous distributions $\mathbb{P}$ and $\mathbb{Q}$ with densities $p(x)$ and $q(x)$. We define $\s{p}$ and $\s{q}$ as a convolution
\begin{align}
&\s{p}(y) = \int  k(y-x)p(x)dx = \conv{k}{p}(y), \\ &\s{q}(y) = \int k(y-x)q(x)dx = \conv{k}{q}(y).
\end{align}
Since $k{>}0$, $\s{p}$ and $\s{q}$ have common support $\mathbb{R}$. 

Since all densities $p(x)$ are absolutely integrable, the Fourier transforms $\F{p}$ and $\F{q}$ exist.
Assuming $\F{k}$ also exists, we can then use the convolution theorem to write
\beq
\F{\s{p}} = \F{k}\F{p}, \hspace{3mm} \F{\s{q}} = \F{k}\F{q}.
\eeq
%
Let\footnote{In fact the more general proof in \appref{app:general_proof} shows that only the weaker condition that $\mathcal{F}\{k\}\neq 0 $ on a countable set is needed.} $\mathcal{F}\{k\}\neq 0$. Then
\begin{align}
\mathcal{F}\{k\}\mathcal{F}\{p\}=\mathcal{F}\{k\}\mathcal{F}\{q\} \Rightarrow \mathcal{F}\{p\}=\mathcal{F}\{q\}.
\end{align}
Using this we can show that the stationary Spread Divergence is valid. A sketch of the proof is as follows:
\begin{align}
\sdiv{p}{q} = 0 &\Leftrightarrow\div{\s{p}}{\s{q}} = 0\\ &\Leftrightarrow \s{p} = \s{q} \; a.e. \\ &\Leftrightarrow \F{k}\F{p} = \F{k}\F{q}\\
& \Leftrightarrow  \F{p} = \F{q}\label{eq:nonzero} \\  & \Leftrightarrow  p = q \; a.e.\Leftrightarrow \mathbb{P}=\mathbb{Q},
\end{align}
where we used the invertibility of the Fourier transform. 

Hence we can define a valid Spread Divergence provided (i) $k(x)$ is a positive probability density function with support $\mathbb{R}$ and (ii) $\mathcal{F}\{k\}\neq 0$. 

As an example consider Gaussian additive spread noise $k(x) = \ndist{x}{0}{\sigma^2}$. This has Fourier transform
\beq
    \F{k}(\omega) = e^{-\frac{\sigma^2\omega^2}{2}} >0. \label{eq:gauss_noise}
\eeq
Similarly, for Laplace noise $k(x) = \frac{1}{2b}e^{-\frac{1}{b}\abs{x}}$,
\beq
\F{k}(\omega)=\sqrt{\frac{2}{\pi}}\frac{b^{-1}}{b^{-2}+\omega^2} >0.  \label{eq:laplace_noise}
\eeq
In both cases $k{>}0$ and $\F{k}{>}0$. Additive Gaussian and Laplace noise can therefore be used to define a valid Spread Divergence.  That is, if the divergence between the spreaded distributions is zero, then the original distributions are equal.

\subsubsection{General Stationary Case \label{sec:general:stationary}}

The previous additive noise setting assumed that $X$ is absolutely continuous. In \appref{app:general_proof} we show how to extend this to all distributions, including implicit generative models. We show that adding absolutely continuous noise $K$ to $X$ defines an absolutely continuous random variable $X+K$, even if $X$ is itself not absolutely continuous (which is the case for implicit generative models). Applying the same additive noise process to both $\mathbb{P}$ and $\mathbb{Q}$ then results in absolutely continuous distributions that have densities $\s{p}$ and $\s{q}$ and common support.  We further show that, provided the characteristic function of the additive random noise variable $K$ is non-zero\footnote{See \appref{app:general_proof} for a weaker condition.},  the noise $K$ can be used to define a valid Spread Divergence between any two distributions with the property that  $\sfdiv{\mathbb{P}}{\mathbb{Q}}=0 \Leftrightarrow \mathbb{P}=\mathbb{Q}$.
This non-zero requirement for the characteristic function is analogous to the characteristic condition on kernels such that the Maximum Mean Discrepancy $\text{MMD}(\mathbb{P},\mathbb{Q})=0 \Leftrightarrow \mathbb{P}=\mathbb{Q}$, see \cite{Sriperumbudur:2011:UCK:1953048.2021077,SriperumbudurFGSL2012,gretton2012kernel}.
%

As an illustration, consider the extreme case of two delta distributions $\mathbb{P}$ and $\mathbb{Q}$ with generalised densities
\begin{align}
    p(x)=\delta(x-\mu_p), \quad q(x)=\delta(x-\mu_q).
\end{align}
In this case $\KL{p}{q}$ is not well defined. For stationary Gaussian noise $p(y{\mid}x)=\ndist{y}{x}{\sigma^2}$,
the `spreaded' distributions are:
\begin{align*}
\tilde{p}(y)&=\int \delta(x-\mu_p)\ndist{y}{x}{\sigma^2}dx=\ndist{y}{\mu_p}{\sigma^2},\\
\tilde{q}(y)&=\int \delta(x-\mu_q)\ndist{y}{x}{\sigma^2}dx=\ndist{y}{\mu_q}{\sigma^2}.
\end{align*}
%
For noise variance $\sigma^2=0.5$ this gives:
\beq
\KL{\tilde{p}}{\tilde{q}}=||\mu_p-\mu_q||_2^2.
\eeq 
Hence $\KL{\tilde{p}}{\tilde{q}}=0 \Leftrightarrow \mathbb{P}=\mathbb{Q}$. It is also worth noting that the spread KL divergence in this case is equal to the squared 2-Wasserstein distance \citep{peyre2019computational,gelbrich1990formula}. Treating $\mu_q$ as a variable, \figref{fig:intro_toy} illustrates the spread KL divergence converging to 0 as $\mu_q$ tends to $\mu_p=0$. 

This treatment of generalised densities allows us to define a divergence for implicit generative models and, by extension, an associated training algorithm, as we describe below.



\subsection{Spread Maximum Likelihood Estimation\label{sec:spread:mle}}
\label{sec:noise:prop}
In \secref{sec:model:not:enough} we noted that in the context of fitting an implicit generative model $\mathbb{Q}$
to training data, simply adding noise to the model distribution $\mathbb{Q}$ and using maximum likelihood does not result in a consistent estimator. In the Spread Divergence case, for data $x_1,\ldots,x_N$ we define the empirical (generalised) density as
\beq
p(x) = \frac{1}{N}\sum_{n=1}^N\delta\br{x-x_n}.
\eeq
For spread noise $p(y|x)$ we then spread both the model $\mathbb{Q}$ and empirical density using
\beq
\tilde{q}(y)= \int p(y|x)q(x)dx, \hspace{2mm} \tilde{p}(y)= \int p(y|x)p(x)dx.
\eeq
We then define the spread log likelihood using
\beq
\tilde{\sett{L}} \equiv \int \tilde{p}(y)\log \tilde{q}(y) dy.
\label{eq:spread:ll}
\eeq
We consider that the data $x_1,\ldots,x_N$ is generated from the same underlying parametric implicit distribution as the model $\mathbb{Q}$ 
\beq
m(x;\theta_q)=\int \delta(x-g_{\theta_q}(z))q(z)dz,
\eeq
but with a different parameters $\theta_p$. Then as $N\rightarrow\infty$ (using the law of large numbers) the spread log likelihood becomes
\beq
\lim_{N\rightarrow\infty}\sett{L} = \int \tilde{m}(y;\theta_p)\log \tilde{m}(y;\theta_q) dy,
\label{eq:av:spread:ll}
\eeq
where 
\beq
\tilde{m}(y;\theta) = \int p(y|x)m(x;\theta)dx.
\eeq
Up to an additive constant this is 
$-\KL{\tilde{m}(y;\theta_p)}{\tilde{m}(y;\theta_q)}$.
%
The spread log likelihood \eqref{eq:av:spread:ll} therefore has a maximum when the spread KL divergence has a minimum. This occurs when the two spread distributions match $\tilde{m}(y;\theta_p) = \tilde{m}(y;\theta_q)$. Furthermore, by the property of the Spread Divergence, this means that the spread log likelihood has a maximum when the distributions match $m(y;\theta_p) = m(y;\theta_q)$, which occurs when $\theta_q=\theta_p$. Hence, the spread log likelihood defines a consistent estimator.  In practice we typically cannot carry out the integral in \eqref{eq:spread:ll} exactly, and resort to a sample approximation, sampling $L$ noisy versions $y_{n,l}$, $l=1,\ldots, L$ of each datapoint $x_n$ to give
\beq
\tilde{\sett{L}} \approx \frac{1}{LN} \sum_{n=1}^N\sum_{l=1}^L\log \tilde{q}(y_{n,l}).
\eeq


The Maximum Likelihood Estimator (MLE) is a cherished approach due to its consistency (convergence to the true parameters in the large data limit) and asymptotic efficiency (achieves the Cramér-Rao lower bound on the variance of any unbiased estimator) - see \citet{casella2002statistical} for an introduction. An interesting question for future study is whether these properties also carry over to the spread MLE. In \appref{app:statistical_property}, we demonstrate that spread MLE (for a certain family of spread noise) has weaker sufficient conditions than MLE for both consistency and asymptotic efficiency. Furthermore, a sufficient condition for the existence of the MLE is that the likelihood function is continuous over a compact parameter space $\Theta$. We provide an example in \appref{app:mle:not:defined} where the maximum likelihood estimator may not exist (since it violates the compactness requirement), but the spread maximum likelihood estimator still exists.

\section{Maximising Discriminatory Power \label{sec:max:disc:power}}

Intuitively, spreading out distributions makes them more similar. More formally, from the data processing inequality (see \appref{app:dpi}), using spread noise will always decrease the $f$-divergence $\fdiv{\s{p}}{\s{q}}\leq \fdiv{p}{q}$ (when $\fdiv{p}{q}$ is well defined). If we use spread MLE to train a model, too much noise may make the spreaded empirical and spreaded model distributions so similar that it becomes difficult to numerically distinguish them. It is useful therefore to learn spread noise $p_\psi(y|x)$ (parameterised by $\psi$) to maximally discriminate between the distributions $\max_{\psi}\div{\tilde{p}}{\tilde{q}}$. In general we need to constrain the spread noise to ensure that the divergence remains bounded.  
The learned noise will discourage overlap between the two spreaded distributions. 

We discuss below two complementary approaches to adjust $p_\psi(y|x)$. 
The first adjusts the covariance for Gaussian $p(y|x)$ and the second uses a mean transformation. In principle, both approaches can be combined and easily generalised to other noise distributions, such as the Laplace distribution. In \secref{sec:delta:vae}, we empirically investigate the benefit of these approaches when scaling the application of Spread Divergence to complex high dimensional problems.

\subsection{Learning the Gaussian Noise Covariance \label{sec:learn_cov}}
In learning Gaussian stationary spread noise $p(y|x)=\ndist{y}{x}{\Sigma}$, the number of parameters in the covariance matrix $\Sigma$ scales quadratically with the data dimension $D$. We therefore define $\Sigma = \sigma^2 I + UU\trans$
where $\sigma^2>0$ is fixed (to ensure bounded Spread Divergence) and $U$ is a learnable $D\times R$ matrix with $R\ll D$.  

\begin{figure}[t]
\begin{center}
\scalebox{0.8}{
\begin{tikzpicture}
\draw (0,0) -- (0,1);
\draw[thick,dotted] (-2,0) -- (2,0);
\draw[thick,dotted] (-2,2) -- (2,2);
\draw (0,0)--(0,0) node[anchor=south east] {$a$};
\draw (0,1)--(0,2) node[anchor=south east] {$b$};
\draw[thick,->,red] (0,0)--(1,0) node[anchor=north] {$A$};
\draw[fill] (0,0) circle (0.5mm); \draw[fill] (0,2) circle (0.5mm);
\end{tikzpicture}
\hspace{1.5cm}
\begin{tikzpicture}
\draw (0,0) -- (0,1);
\draw[thick,dotted] (-2,0) -- (2,0);
\draw[thick,dotted] (-2,2) -- (2,2);
\draw (0,0)--(0,0) node[anchor=south east] {$a$};
\draw (0,1)--(0,2) node[anchor=south east] {$b$};
\draw[thick,->,red] (0,0)--(1,0) node[anchor=north] {$A$};
\draw[fill] (0,0) circle (0.5mm); \draw[fill] (0,2) circle (0.5mm);
\draw (0,0) ellipse (2.2cm and 0.5cm);
\draw (0,2) ellipse (2.2cm and 0.5cm);
\draw[thick,->] (0,1)--(1,1) node[anchor=north] {$u$};
\end{tikzpicture}}
\end{center}
\caption{Left: The lower dotted line denotes Gaussian distributed data $p(x)$ with support only along the linear subspace defined by the origin $a$ and direction $A$. The upper dotted line denotes Gaussian distributed data $q(x)$ with support different from $p(x)$. Optimally, to maximise the Spread Divergence between the two distributions, for fixed noise entropy, we should add noise that preferentially spreads out along the directions defined by $p$ and $q$, as denoted by the ellipses. \label{fig:spread}}
\end{figure}
As a simple example that can be computed exactly, we consider a implicit generative models with generalised densities $p$ and $q$ that generate data in separated linear subspaces, 
\begin{align}
p(x)&=\int \delta\br{x - a - Az}p(z)dz\\
q(x)&=\int \delta\br{x-b - Bz}p(z)dz,
\end{align}
with $p(z)=\ndist{z}{0}{I_Z}$. The spreaded densities are then 
\[
\s{p}(y) = \ndist{y}{a}{AA\trans + \Sigma},\hspace{2mm} \s{q}(y) = \ndist{y}{b}{BB\trans + \Sigma}.
\]
As $\Sigma$ tends to zero,  $\KL{\s{p}}{\s{q}}$ tends to infinity. We therefore constrain $\Sigma = \sigma^2 I + u u\trans$, where $\sigma^2$ is fixed and $u\trans u=1$. 
Also, for calculational simplicity, we assume $A=B$. 
The Spread Divergence $\KL{\s{p}}{\s{q}}$ is then maximised for the noise direction $u$ pointing orthogonal to the vector $\br{AA\trans+\sigma^2I}^{-1}\br{b-a}$. 
The noise thus concentrates along directions defined by $p$ and $q$, see \figref{fig:spread}. 

\subsection{Learning a Mean Transformation \label{sec:injective}}

Consider spread noise $p(y{\mid}x) = k(y-f(x))$ for injective\footnote{Since the co-domain of $f$ is determined by its range, injective indicates invertible in this case.} $f$ and stationary $k$.  Then, we define
\beq
\s{p}(y) = \int k(y-f(x))p_x(x)dx.
\label{eq:injective}
\eeq
Using a change of variables $s=f(x)$
\begin{align}
&\s{p}(y) = \int k(y-s)p_s(s)dz,\label{eq:inf:sd}\\ 
&p_s(s) = p_x(f^{-1}(s))/J\br{x=f^{-1}(s)},
\end{align}
where $J$ is the absolute Jacobian of $f$. This is a valid Spread Divergence since \eqref{eq:inf:sd} is in the form of standard stationary spread noise, but on an invertible transformed variable $s$.  
Hence, $\div{\s{p}_y}{\s{q}_y}=0  \Leftrightarrow p_s=q_s \Leftrightarrow p_x=q_x$. 
Each injective $f$ gives a different noise $p(y|x)$ and hence we can search for the best noise implicitly by learning $f$.

\section{Applications \label{sec:apps}}

As an application of the spread MLE, we use it to train implicit generative models with generalised density
\beq
p_\theta(x) = \int  \delta\br{x-g_\theta(z)}p(z) dz,
\label{eq:implicit:model}
\eeq
where $\theta$ are the parameters of the encoder $g$.  We show that, despite the likelihood gradient not being available, we can nevertheless successfully train such models using slightly modified versions of standard likelihood based training approaches, such as variational algorithms \citep{Barber:2012:BRM:2207809}. In particular we discuss learning a low dimensional linear ICA model and a high dimensional deep generative model. 


\subsection{Implicit Linear Models: Deterministic ICA \label{sec:ica}}

ICA (Independent Components Analysis) corresponds to the model $p(x,z)=p(x{\mid}z) \prod_i p(z_i)$, where the independent components $z_i$ follow a non-Gaussian distribution. For Gaussian noise ICA an observation $x$ is assumed to be generated by the process $p(x{\mid}z) = \prod_j\ndist{x_j}{g_j(z)}{\gamma^2}$
where $g_i(z)$ mixes the independent latent process $z$. In linear ICA, $g_j(z)=a_j\trans z$ where $a_j$ is the $j^{th}$ column on the mixing matrix $A$. For zero observation noise $\gamma^2=0$ this corresponds to a linear implicit generative model. For training data $x_1,\ldots,x_N$ a standard maximum likelihood approach to learning $A$ is to maximise
\beq
\ave{\log p(x)}{\hat{p}(x)},
\eeq
where $p(x)$ is the marginal of the joint $p(x,z)$ and we define the empirical distribution
\beq
\hat{p}(x)=\frac{1}{N}\sum_{n=1}^N\delta\br{x-x_n}.
\eeq
Since this is a latent variable model, it is natural to apply the EM algorithm \cite{dempster1977maximum} to learn $A$. However, for zero, or even small observation noise $\gamma^2$, it is well known that EM is ineffective \citep{Bermond99approximatelikelihood,WINTHER2007858}. To see this, consider $\sz{X}=\sz{Z}$ (where $\sz{X}$ and $\sz{Z}$ are the dimension of the data and latents respectively) and invertible $A$. At iteration $k$ the EM algorithm has an estimate $A_k$ of the mixing matrix. It is straightforward to show that the M-step updates $A_k$ to
\beq
A_{k+1} = \ave{x z\trans}{}\ave{zz\trans}{}^{-1},
\label{eq:Aupdate}
\eeq
where, for zero observation noise ($\gamma=0$),
\begin{align}
&\ave{xz\trans}{} = \frac{1}{N}\sum_n x_n\br{A_k^{-1}x_n\trans} =
\hat{S} A_k\ntrans, \\&\ave{zz\trans}{} = A_k^{-1}\hat{S}A_k\ntrans,
\end{align}
and $\hat{S}\equiv \frac{1}{N}\sum_n x_nx_n\trans$ is the moment matrix of the data. Thus,
$A_{k+1} = \hat{S}A_k\ntrans\br{A_k^{-1}\hat{S}A_k\ntrans}^{-1}= A_{k}$
and the algorithm `freezes'. Similarly, for low noise $\gamma\ll 1$, progress critically slows down. Thus trying to train a deterministic ICA model by simply adding noise to the model will not help directly (see also \secref{sec:model:not:enough}).
\begin{figure*}[t]
\vspace{-3mm}
	\centering
		\subfloat[Error versus observation noise $\gamma$.]{\includegraphics[width=0.36\linewidth]{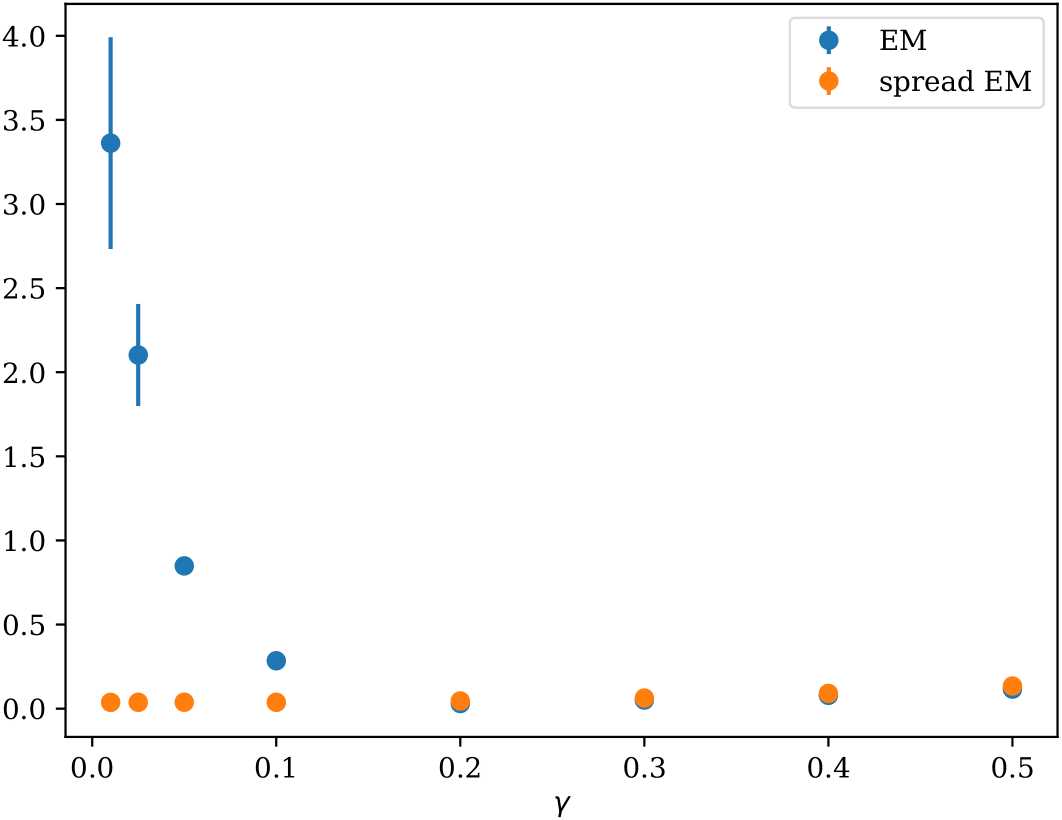}}
 		\hspace{5mm}
		\subfloat[Error versus number of training points.]{\includegraphics[width=0.36 \linewidth]{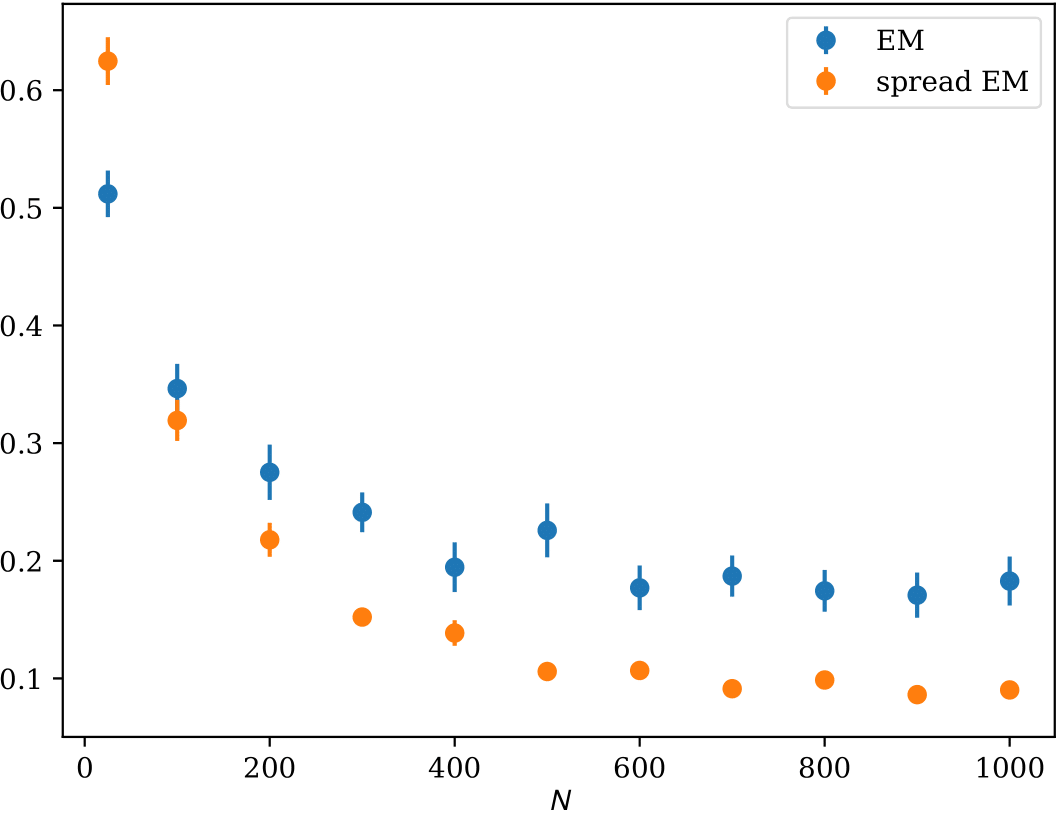}}
	\caption{
		Relative error $|A^{est}_{ij}-A^{true}_{ij}|/|A^{true}_{ij}|$ versus observation noise (a) and number of training points (b).
		(a) For $\sz{X}=20$ dimensional observations and $\sz{Z}{=}10$ dimensional latent variables, we generate $N{=}20000$ data points from the model $x=Az+\gamma\sett{N}(0,I_X)$, for independent zero mean unit variance Laplace $z$. The elements of $A$ are uniform random $\pm 1$. We use $S_y{=}1$, $S_z{=}1000$ samples and 2000 EM iterations to estimate $A$. The error is averaged over all $i,j$ and 10 experiments. We also plot standard errors around the mean relative error. In blue we show the error in learning $A$ using the standard EM algorithm. As $\gamma{\rightarrow}0$, the error blows up as the EM algorithm `freezes'. In orange we plot the error for spread noise EM; no slowing down occurs as the observation noise $\gamma$ decreases. 
		In (b), apart from small $N$, the spread EM algorithm error is lower than for the standard EM algorithm.  Here $\sz{Z}{=}5$, $\sz{X}{=}10$, $S_y{=}1$, $S_z{=}1000$, $\gamma=0.2$, with 500 EM updates used. Results are averaged over 50 runs of randomly drawn $A$. \label{fig:ica}}
\end{figure*}

To deal with small noise and the limiting case of a deterministic model ($\gamma=0$), we consider Gaussian spread noise $p(y{\mid}x)=\ndist{y}{x}{\sigma^2 I_X}$ to give
\begin{align}
&p(y,z) = \int p(y{\mid}x)p(x,z) dx \\&\hspace{3mm}= \prod_j\ndist{y_j}{g_j(z)}{\br{\gamma^2+\sigma^2} I_X} \prod_i p(z_i).
\end{align}
Using spread noise, the empirical distribution is replaced by the spreaded empirical distribution
\beq
\hat{p}(y)=\frac{1}{N}\sum_n \ndist{y}{x^n}{\sigma^2I_X}.
\eeq
We then learn the model parameters by maximising the spread log likelihood (see \secref{sec:spread:mle})
\beq
\ave{\log p(y)}{\hat{p}(y)}. \label{eq:sp:mle}
\eeq
Since this is of the form of a latent variable model, we can use an EM algorithm to maximise \eqref{eq:sp:mle}. 
The M-step to maximise the spread log likelihood  has the same form as \eqref{eq:Aupdate} but with modified statistics
\begin{align}
&\ave{yz\trans}{} = \frac{1}{N}\sum_n \int p(y,z|n) yz\trans dzdy,\label{eq:yz}\\
&\ave{zz\trans}{} = \frac{1}{N}\sum_n \int p(y,z|n) zz\trans dzdy.
\end{align}
where
\beq
p(y,z|n)\equiv \ndist{y}{x^n}{\sigma^2}p(z{\mid}y)
\eeq
with
\begin{align}
   p(z{\mid}y) &= \frac{1}{Z_q(y)}\ndist{z}{\mu(y)}{\Sigma}\prod_i p(z_i),\\
Z_q(y) &= \int \ndist{z}{\mu(y)}{\Sigma}\prod_i p(z_i) dz,
\end{align}
Here $Z_q(y)$ is a normaliser and
\beq
\Sigma = (\gamma^2+\sigma^2)\br{A\trans A}^{-1}, \hspace{3mm} \mu(y)  = \br{A\trans A}^{-1}Ay.
\eeq
%
%
Since the posterior $p(z{\mid}y)$ peaks around $\ndist{z}{\mu(y)}{\Sigma}$, we rewrite \eqref{eq:yz} as
\begin{align}
\ave{yz\trans}{}&=\frac{1}{N}\sum_n \int \ndist{y}{x^n}{\sigma^2I_X}\ndist{z}{\mu(y)}{\Sigma}\nonumber\\&\hspace{20mm}\times\frac{\prod_i p(z_i)}{Z_q(y)}yz\trans dzdy
\end{align}
and similarly for  $\ave{zz\trans}{}$.
Writing the expectations with respect to $\ndist{z}{\mu(y)}{\Sigma}$ allows for a simple but effective importance sampling approximation focused on regions of high probability.
We implement this update by drawing $S_y$ samples from $\ndist{y}{x_n}{\sigma^2I_X}$ and, for each $y$ sample, we draw $S_z$ samples from $\ndist{z}{\mu(y)}{\Sigma}$.  This scheme has the advantage over more standard variational approaches, see for example \cite{WINTHER2007858},  in that we obtain a consistent estimator of the M-step update for $A$.

We show results for a toy experiment in \figref{fig:ica}, learning the mixing matrix $A$ in a deterministic non-square setting. Note that standard algorithms such as FastICA \citep{fastica} fail in this case. The spread noise is set to $\sigma=\max(0.001,2.5*\text{sqrt}(\text{mean}(AA\trans)))$. This modified EM algorithm thus learns a good approximation of $A$, with no critical slowing down. 


\subsection{Non-linear Implicit Models: $\delta$-VAE\label{sec:delta:vae}}

A deep implicit generative model has generalised density
\beq
p_\theta(x){=}\int \delta\br{x-g_\theta(z)}p(z)dz,
\eeq
where $g_\theta$ is a deep neural network, see for example \cite{goodfellow2014generative}. As discussed, for $\sz{Z}<\sz{X}$ we cannot use standard maximum likelihood approaches to train this model. 
%
%
To address this, we consider using the spread MLE approach \secref{sec:spread:mle}. For training data $x_1,\ldots, x_N$ the empirical distribution is
\beq
\hat{p}(x) = \frac{1}{N}\sum_{n=1}^N\delta\br{x-x_n}.
\eeq
For Gaussian spread noise $p(y|x){=}\ndist{y}{x}{\sigma^2I_X}$, the spreaded empirical distribution is
\beq
\s{p}(y) = \frac{1}{N}\sum_{n=1}^N \ndist{y}{x_n}{\sigma^2 I_X},
\eeq
and the spreaded model is
\begin{align}
\s{p}_\theta(y) &= \int  \ndist{y}{g_\theta(z)}{\sigma^2I_X}p(z) dz\\ &= \int  p_\theta(y{\mid}z)p(z) dz.
\end{align}
We then maximise the spread log likelihood
\beq
\int \s{p}(y) \log \s{p}_\theta(y)dy.
\eeq
Typically, the integral over $y$ is intractable, in which case we resort to a sampling estimation
\beq
\frac{1}{NS}\sum_{n=1}^N\sum_{s=1}^S \log \s{p}_\theta(y^n_s),
\eeq
where $y^n_s$ is a sample from $p(y_n|x_n)=\ndist{y^n}{x_n}{\sigma^2I_X}$.

For non-linear $g$, the distribution $\s{p}_\theta(y)$ is usually intractable and we therefore use the variational lower bound
\begin{align}
\log \s{p}_\theta(y) &\geq \int q_\phi(z{\mid}y)(-\log q_\phi(z{\mid}y)\nonumber\\&\hspace{15mm} + \log\br{p_\theta(y{\mid}z)p(z)}) dz.
\end{align}
This approach is a straightforward extension of the standard variational autoencoder (VAE) method \cite{vae} and in \appref{app:nn} we derive a lower variance objective and detail the learning procedure.  We dub this model and associated Spread Divergence training the `$\delta$-VAE'. 

For learning the spread noise we use the approach outlined in \secref{sec:max:disc:power}. To learn the mean transformation function, we used an invertible residual network \citep{behrmann2018invertible} $f_\psi:\mathbb{R}^D\rightarrow\mathbb{R}^D$ where $f_\psi=(f_\psi^1\circ\ldots\circ f_\psi^T)$ denotes a ResNet with blocks $f_\psi^t=I(\cdot)+g_{\psi_t}(\cdot)$. Then $f_\psi$ is invertible if the Lipschitz-constants $Lip(g_{\psi_t}){<}1$, for all $t\in\cb{1,\ldots,T}$. Note that when using the Spread Divergence for training we only need samples from $\s{p}(y)$ which can be obtained from \eqref{eq:injective} by first sampling $x$ from $p_x(x)$ and then $y$ from $p(y|x)=k(y{-}f(x))$; this does not require computing the Jacobian or inverse ${f_\psi}^{-1}$.


\begin{figure}[t]
    \centering
    \subfloat[Fixed Laplace Spread Noise]{
    \includegraphics[width=0.9\linewidth]{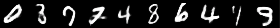}}
    \vskip 1 mm
    \subfloat[Fixed Gaussian Spread Noise]{
    \includegraphics[width=0.9\linewidth]{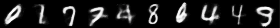}}
    \vskip 1 mm
    \subfloat[Learned Gaussian Spread Noise]{
    \includegraphics[width=0.9\linewidth]{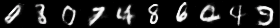}}
  \caption{Samples from a deep implicit generative model trained using $\delta$-VAE with (a) Laplace spread noise with fixed covariance, (b) Gaussian spread noise with fixed covariance and (c)  Gaussian spread noise with learned covariance. We first train with one epoch a standard VAE as initialization to all models and keep the latent code $z\sim \ndist{z}{0}{I_Z}$ fixed when sampling from these models thereafter, so we can more easily compare the sample quality. See also \figref{fig:app:mnist} for further samples.
    \label{fig:mnist}}
\end{figure}

\begin{figure}[t]
    \vspace{-3.5mm}
    \centering
    \subfloat[$\delta$ Fixed spread noise]{
        \includegraphics[width=0.47\linewidth,valign=c]{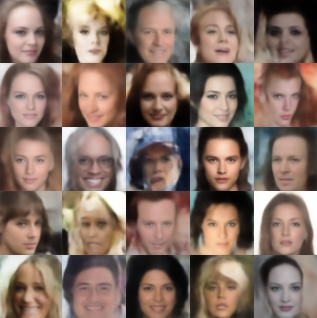}
        }
    \subfloat[$\delta$ Learned spread noise]{
        \includegraphics[width=0.47\linewidth,valign=c]{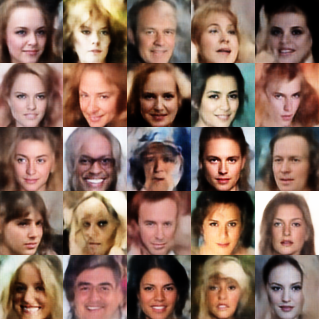}
        }
    \caption[]{Samples from a deep implicit generative model trained using $\delta$-VAE with (a) fixed and (b) learned spread with the mean transformation method. See also \figref{fig:more:samples:celeb} for more samples. We use a similar sampling strategy as in the MNIST experiment to facilitate sample comparison between the different models -- see \appref{app:celeba}. \label{fig:celeba}}
\end{figure}

\textbf{MNIST Experiment:} We trained a $\delta$-VAE on MNIST \citep{lecun-mnisthandwrittendigit-2010} with (i) fixed Laplace spread noise, as in \eqref{eq:laplace_noise}, (ii) fixed Gaussian spread noise, as in \eqref{eq:gauss_noise} and (iii) Gaussian noise with learned covariance, as in \secref{sec:learn_cov}, with rank  $R=20$; $g_\theta(\cdot)$ is a neural network that contains 3 feed forward layers. See \appref{app:mnist} for further details. 

Figures \ref{fig:mnist}(a,b,c) show samples from $p_\theta(x)$ for these models. MNIST is a relatively easy problem in the sense that it is hard to distinguish between the quality of the fixed and learned noise samples.  We speculate that Laplace noise appears to improve image sharpness since this noise focuses attention on discriminating between points close to the data manifold (since the Laplace distribution is leptokurtic and has a higher probability of generating points close to the data manifold than the Gaussian distribution). 



\textbf{CelebA Experiment:}  We trained a $\delta$-VAE on the CelebA dataset \citep{liu2015faceattributes} with (i) fixed and (ii) learned spread using the mean transformation method as discussed in \secref{sec:injective}. We compare to results from a standard VAE with fixed Gaussian noise $p(x|z)=\ndist{x}{g_\theta(z)}{0.5I_X}$ \citep{tolstikhin2017wasserstein}, where $g_\theta(\cdot)$ is a neural network contains 4 convolution layers. 

For (i) the fixed Spread Divergence uses Gaussian noise $\ndist{y}{x}{0.25I_X}$. For (ii) we use Gaussian noise with a learned injective function in the form of a ResNet: $f_\psi(\cdot)=I(\cdot)+g_\psi(\cdot)$ - see \appref{app:celeba} for details. 
Figure \ref{fig:celeba} shows samples from our $\delta$-VAE for (i) and (ii) (with $g_\theta(z)$ initialised to the fixed-noise setting).  It is notable how the `sharpness' of the image samples substantially increases when learning the spread noise. Table \ref{table:fid} shows Frechet Inception Distance (FID) \citep{heusel2017gans} score comparisons between different baseline algorithms for implicit generative model training\footnote{FID scores were computed using \url{github.com/bioinf-jku/TTUR} based on 10000 samples.}. 
The $\delta$-VAE significantly improves on the standard VAE result. Learning the mean transformation improves on the fixed-noise $\delta$-VAE. Indeed the mean transformation $\delta$-VAE results are comparable to popular GAN and WAE models \citep{gulrajani2017improved,berthelot2017began,arjovsky2017wasserstein,kodali2017convergence,mao2017least,fedus2017many,tolstikhin2017wasserstein}. Whilst the $\delta$-VAE results are not state-of-the-art, we believe it is the first time that implicit models have been trained using a principled maximum likelihood based approach. 
By increasing the complexity of the generative model $g_\theta$ and injective function $f_\psi$, or using better choices of noise, the results may become more competitive with state-of-the-art GAN models\footnote{We also tried learning the image generator using Laplace spread noise. However, the colour of the sampled images becomes overly intense and we leave it to future work to address this.}.


\begin{table}
\caption{CelebA FID Scores. The $\delta$-VAE results are the average over 5 independent measurements. The scores of the GAN models are based on a large-scale hyperparameter search and take the best FID obtained \citep{lucic2018gans}. The results of the VAE model and both WAE-based models are from \cite{tolstikhin2017wasserstein}.\label{table:fid}}
\scalebox{0.85}{\begin{tabular}{c c | c c }
\toprule
    \cmidrule(r){1-2}
    Encoder-Decoder Models      & FID  & GAN Models  & FID  \\
    \midrule
    VAE &  63.0  & WGAN GP & 30.0   \\
    \textbf{$\delta$-VAE with fixed spread}    & \textbf{52.7}   & BEGAN & 38.9     \\
    \textbf{$\delta$-VAE with learned spread}    & \textbf{46.5}  & WGAN  & 41.3   \\
    \cline{1-2}
    & & DRAGAN  & 42.3\\
    WAE-MMD     & 55.0  &  LSGAN & 53.9\\
    WAE-GAN & 42.0  &  NS GAN & 55.0\\
    && MM GAN &  65.6\\
    \bottomrule
\end{tabular}}
\end{table}

\section{Related Work \label{sec:related_work}}


\textbf{Instance noise:}
The instance noise trick to stabilize GAN training \citep{roth2017stabilizing,sonderby2016amortised} is a special case of Spread Divergence using fixed Gaussian noise. Whilst other similar tricks, e.g. \cite{furmston-barber-09}, have been proposed previously, we believe it is important to state the more general utility of the spread noise approach.

\textbf{$\delta$-VAE versus WAE}:
The Wasserstein Auto-Encoder \citep{tolstikhin2017wasserstein} is another implicit generative model that uses an encoder-decoder architecture. The major difference to our work is that the $\delta$-VAE is based on the KL divergence, which corresponds to MLE, whereas the WAE uses the Wasserstein distance.

\textbf{$\delta$-VAE versus denoising VAE}: The denoising VAE \citep{im2017denoising} uses a VAE with noise added to the data only.  In contrast, for our $\delta$-VAE, spread MLE adds noise to both the data and the model. Therefore, the denoising VAE cannot recover the true data distribution, whereas in principle the $\delta$-VAE with spread MLE can. 

\textbf{MMD GAN with kernel learning}:
Learning a kernel to increase discrimination is also used in MMD GAN \citep{li2017mmd}. Similar to ours, the kernel in MMD GAN is constructed by $\tilde{k}=k\circ f_\psi$, where $k$ is a fixed kernel and $f_\psi$ is a neural network. To ensure the MMD distance $M_{k\circ f_\psi}(p,q)=0\Leftrightarrow p=q$, this requires $f_\psi$ to be injective \citep{gretton2012kernel}. However, in this framework, $f_\psi(x)$ usually maps $x$ to a lower dimensional space. This is crucial for MMD because the amount of data required to produce a reliable estimator grows with the data dimension \citep{ramdas2015decreasing} and the computational cost of MMD  scales quadratically with the amount of data. Whilst using a lower-dimensional mapping makes MMD more practical it also makes it difficult to construct an injective function $f$. For this reason, heuristics such as the auto-encoder regularizer \citep{li2017mmd} are considered. In contrast, for the $\delta$-VAE with spread MLE, the cost of estimating the divergence is linear in the number of data points. Therefore, there is no need for $f_\psi$ to be a lower-dimensional mapping; guaranteeing that $f_\psi$ is injective is straightforward for the $\delta$-VAE.

\textbf{Flow-based generative models}:
Invertible flow-based functions \citep{rezende2015variational} have been used to boost the representation power of generative models. Our use of injective functions is quite distinct from the use of flow-based functions to boost generative model capacity. In our case, the injective function $f$ does not change the model - it only changes the divergence. For this reason, the Spread Divergence doesn't require the log determinant of the Jacobian, which is required in \cite{rezende2015variational,behrmann2018invertible}, meaning that more general invertible functions can be used to boost the discriminatory power of a Spread Divergence.

\section{Summary}

We described how to define a divergence even when two distributions don't have valid probability density functions or have the same support. 
%
We showed that defining divergences this way enables us to train implicit generative models using standard likelihood based approaches.
%
%
%
In principle, we can learn the true data generating implicit generative model by using any valid Spread Divergence. 
In practice, however, the quality of the learned model can depend strongly on the choice of spread noise. We therefore investigated learning spread noise to maximally discriminate between two distributions. 
We found that the resulting training approach is numerically stable and that it can significantly improve the quality of the learned model. 
%

There are several directions for further investigation: (i) the broader family of spread noise and their properties, including statistical efficiency; (ii) optimal noise selection for different tasks;
(iii) the connections between Spread Divergence and other distance (or divergence) measure, e.g. MMD, Wasserstein Distance. 
\subsubsection*{Acknowledgments}
We would like to thank a reviewer for improving our consideration of non-absolutely continuous distributions and Li Zhu for useful discussions. \newpage
\bibliography{icml_conference}
\bibliographystyle{icml2020}
\newpage
\appendix

\onecolumn

\section{Annealing the Noise\label{sec:annealing:noise}}

In \secref{sec:model:not:enough} we discussed the common approach to first adding noise to a model $\mathbb{Q}$ in order to define a proper density and then using maximum likelihood to fit that `noised model' to data. 

We can use standard Woodberry identities to rewrite the expected log likelihood \eqref{eq:log:lik:limit} as
\beq
-\frac{\theta_p^2}{\sigma^2} + \frac{ \br{\theta_p\trans\theta_q}^2}{\sigma^2\br{\sigma^2+\theta_q^2}} - \log\br{1+\frac{\theta_q^2}{\sigma^2}}-D\log \sigma^2.
\label{eq:wood}
\eeq
where $D=\sz{\theta_p}$.

Differentiating wrt $\theta_q$, we note that the optimal solution is given when
\beq
\theta_q=\gamma\theta_p,
\eeq
for scalar $\gamma$. Plugging this form back into \eqref{eq:wood} we find that the optimum is obtained when
\beq
\theta_q = \sqrt{\frac{\theta_p^2-\sigma^2}{\theta_p^2}}\theta_p.
\eeq

For finite Gaussian noise $\sigma^2 > 0$ the resulting estimator for the toy model in \secref{sec:model:not:enough} is therefore not consistent. 

\begin{figure}[H]
\includegraphics[width=0.6\textwidth]{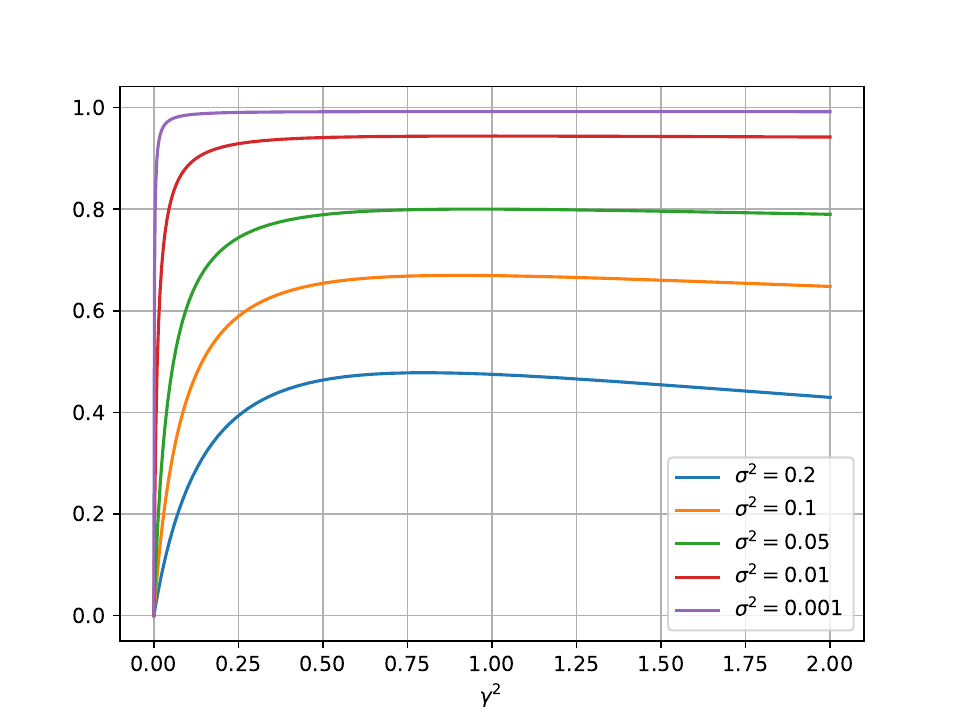}
\caption{The (modified) expected log likelihood \eqref{eq:scaled:exp:log:lik} when adding noise $\sigma^2$ to the model only and for unit length true data generating parameter $\theta_p^2=1$. The $x$-axis is the value $\gamma^2$ assuming that the optimal $\theta_q$ is of the form $\theta_q=\gamma\theta_p$. As we see, as $\sigma^2\rightarrow 0$ the scaled objective becomes flat around the optimum point $\gamma^2=1$.\label{fig:scaling}}
\end{figure}

A natural question is what would happen if one uses a numerical optimisation of \eqref{eq:wood} but anneals the noise $\sigma^2$ to zero during the optimisation process?  As $\sigma^2$ tends to zero, the expression \eqref{eq:wood} blows up. This means that a naive approach to annealing $\sigma^2$ towards zero whilst using a standard optimisation technique is unlikely to result in $\theta_q$ converging to $\theta_p$. However, if one considers removing the additive constant $D\log\sigma^2$ and multiplying the remaining objective by $\sigma^2$, the resulting quantity
\beq
\frac{ \br{\theta_p\trans\theta_q}^2}{\br{\sigma^2+\theta_q^2}} - \sigma^2\log\br{1+\frac{\theta_q^2}{\sigma^2}},
\label{eq:scaled:exp:log:lik}
\eeq
is well-behaved as $\sigma^2\rightarrow 0$, as plotted in \figref{fig:scaling}. 

Nevertheless, the objective \eqref{eq:scaled:exp:log:lik} becomes flat (with respect to $\theta_q$) around the optimum as $\sigma^2\rightarrow 0$.  In \figref{fig:scaling} we plot the scaling behaviour of the objective \eqref{eq:scaled:exp:log:lik}, assuming $\theta_q=\gamma\theta_p$, showing how it becomes flat with respect to $\gamma$ as $\sigma^2$ is annealed towards zero. This means that a standard first-order numerical optimisation approach, even for this modified objective, will result in a `critical slowing down' phenomenon, leading to $\theta_q$ not updating. This might be fixable by taking the curvature of the objective into consideration.

However, addressing all the above issues requires an understanding of the small $\sigma^2$ behaviour of the original objective; dealing with arbitrarily large constants, arbitrarily large scaling and loss of curvature. In general, such insight is unlikely to be available for any given implicit generative model. Thus, we are doubtful that it will be possible to find an annealing schedule and associated general numerical optimisation procedure that will result in a consistent estimator.

\section{Noise Requirements for Discrete Distributions\label{sec:disc: noise:rec}}

Our main interest is to define a new divergence in situations where the original divergence $\div{p}{q}$ is itself not defined. 
For discrete variables $x\in\cb{1,\ldots,n}$, $y\in\cb{1,\ldots,n}$, the spread $P_{ij} = p(y=i{\mid}x=j)$ must be a distribution; $\sum_i P_{ij}=1$, $P_{ij}\geq 0$, and
\begin{small}
\begin{align}
 &\tilde{p}_i\equiv \sum_j P_{ij}p_j = \sum_j P_{ij}q_j\equiv \tilde{q}_i  \hcm \forall i \\ &\hspace{15mm} \Rightarrow \hcm p_j = q_j \hcm\forall j,  \label{eq:inj} 
\end{align}
\end{small}which is equivalent to the requirement that the matrix $P$ is invertible. In addition, for the Spread Divergence to exist in the case of $f$-divergences, $\s{p}$ and $\s{q}$ must have the same support. This requirement is guaranteed if
\begin{align}
\sum_j P_{ij}p_j >0 , \hcm  \sum_j P_{ij}q_j >0  \hcm \forall i,
\end{align}
which is satisfied if $P_{ij}{>}0$. Therefore, in general there is a space of spread distributions $p(y{\mid}x)$ that define a valid Spread Divergence in the discrete case.

\section{Validity of Stationary Spread $f$-Divergence\label{app:general_proof}}
\begin{lemma}\label{lemma:1}
let $X$ and $Y$ be two random variables with Borel probability measure $\mathbb{P}_X$ and $\mathbb{P}_Y$. Let $K$ be an absolutely continuous random variable that is independent of $X$ and $Y$ and has density function $p_K(x)$. We define $\tilde{X}$ and $\tilde{Y}$ as
$$\tilde{X}=X+K,\quad\tilde{Y}=Y+K,$$with distribution $\mathbb{P}_{\tilde{X}}$ and $\mathbb{P}_{\tilde{Y}}$.
Then $\tilde{X}$ and $\tilde{Y}$ are absolutely continuous with density functions
$$p_{\tilde{X}}(\tilde{x})=\int_x  p_K(\tilde{x}-x)d\mathbb{P}_X,\quad p_{\tilde{Y}}(\tilde{y})=\int_y p_K(\tilde{y}-y)d\mathbb{P}_Y.$$
\begin{proof}
The proof  can be found in \citet[Theorem 2.1.16]{durrett2019probability}.
\end{proof}
\end{lemma}

\begin{thm}
Let $X$ and $Y$ be two random variables\footnote{We don't require $X$ (or $Y$) to be absolutely continuous.}  with Borel probability measure $\mathbb{P}_X$ and $\mathbb{P}_Y$. Let the stationary spread noise $K$ be an absolutely continuous random variable that is independent of $X$ and $Y$, and its density function $p_K(x)$ has support\footnote{The extension to higher dimensions is straightforward.} $\mathbb{R}$. Using lemma(\ref{lemma:1}) we define spreaded random variables $\tilde{X}=X+K$, $\tilde{Y}=Y+K$ with density functions $p_{\s{X}}$, $p_{\s{Y}}$.  We then define the stationary  spread $f$-divergence between $\mathbb{P}_{X}$ and $\mathbb{P}_{Y}$ as
$$\sfdiv{\mathbb{P}_X}{\mathbb{P}_Y}\equiv\fdiv{p_{\tilde{X}}}{p_{\tilde{Y}}}\equiv \int f\br{\frac{p_{\tilde{X}}(x)}{p_{\tilde{Y}}(x)}}p_{\tilde{Y}}(x)dx.$$
Furthermore, denote the characteristic function\footnote{When a distribution $\mathbb{P}_X$ allows a density function $p_X$, its characteristic function is equal to the Fourier transform of the density function: $\phi_X=\mathcal{F}\{p_X\}$, so the Fourier transform treatment used in the main text can be seen as a special case of the characteristic function treatment.} of the spread noise $K$ by $\phi_K$. Given $\phi_{K}\neq 0$ or $\phi_{K}>0$ on at most a countable set, then the stationary spread $f$-divergence is a valid divergence with the properties
$$\sfdiv{\mathbb{P}_X}{\mathbb{P}_Y}\geq 0,\quad\sfdiv{\mathbb{P}_X}{\mathbb{P}_Y}=0\Leftrightarrow \mathbb{P}_X=\mathbb{P}_Y.$$
\end{thm}
\begin{proof}

The proof contains the following two steps. 

\textbf{First step:} We show that if $K$ is an absolutely continuous random variable and its density function $p_K$ has support $\mathbb{R}$, then $\sfdiv{\mathbb{P}_X}{\mathbb{P}_Y}=0\Leftrightarrow \mathbb{P}_{\tilde{X}}=\mathbb{P}_{\tilde{Y}}.$
By Lemma \ref{lemma:1}, we have $\s{X}$ and $\s{Y}$ are absolutely continuous and allow probability density functions $p_{\s{X}}$ and $p_{\s{Y}}$. Since $p_K$ has support $\mathbb{R}$, $p_{\s{X}}$ and $p_{\s{Y}}$ will also have support $\mathbb{R}$.
The $f$-divergence between two absolutely continuous distributions with common support is equal to zero if and only if two distributions are equal \citep{fdiv1,csiszar1972class}. We have $\fdiv{p_{\tilde{X}}}{p_{\tilde{Y}}}=0\Leftrightarrow \mathbb{P}_{\tilde{X}}=\mathbb{P}_{\tilde{Y}}.$ Therefore,
$$\sfdiv{\mathbb{P}_X}{\mathbb{P}_Y}=0\Leftrightarrow \mathbb{P}_{\tilde{X}}=\mathbb{P}_{\tilde{Y}}.$$

\textbf{Second step:} We show that if the characteristic function of the spread noise $\phi_K\neq0$ or $\phi_K=0$ on at most a countable set then $\mathbb{P}_{\tilde{X}}=\mathbb{P}_{\tilde{Y}}\Leftrightarrow \mathbb{P}_X=\mathbb{P}_Y.$

The characteristic function of a probability measure $\mathbb{P}_X$ is defined as $\phi_X(w)=\int e^{iwx}d\mathbb{P}_X$.
Since a probability measure is uniquely determined by its characteristic function \citep[Theorem 4.3]{kallenberg2006foundations}, we have
$$ \mathbb{P}_{\tilde{X}}=\mathbb{P}_{\tilde{Y}}\Leftrightarrow \phi_{\tilde{X}}=\phi_{\tilde{Y}}.$$
Using the fact that the characteristic function of the sum of two random variables is equal to the product of their characteristic functions  \citep[Theorem 3.3.2]{durrett2019probability}, we can write
$$\phi_{\tilde{X}}=\phi_{\tilde{Y}}\Leftrightarrow \phi_X\phi_K=\phi_Y\phi_K.$$
When $\phi_K\neq0$, we have
$\phi_X\phi_K=\phi_Y\phi_K\Leftrightarrow \phi_X=\phi_Y$.

When $\phi_K=0$ on at most a countable set $\mathcal{C}$, we show that $\phi_X\phi_K=\phi_Y\phi_K\Leftrightarrow \phi_X=\phi_Y$ still holds. We prove this by contradiction:

We assume there is a point $w_0\in \mathcal{C}$ where $\phi_X(w_0)\neq \phi_Y(w_0)$. Without loss of generality, we assume $\phi_X(w_0)-\phi_Y(w_0)=\delta > 0$. For points $w_0+h$ that are not in $\mathcal{C}$, we have $\phi_K(w_0+h)\neq 0$, so $\phi_X\phi_K=\phi_Y\phi_K$ implies $\phi_X(w_0+h)-\phi_Y(w_0+h)=0$. Since the characteristic function of a distribution is uniform continuous \citep[Theorem 3.3.1]{durrett2019probability}, we have $\delta=\phi_X(w_0+h)-\phi_Y(w_0+h)\rightarrow 0$ when $h\rightarrow 0$, which leads to a contradiction (since $\delta$ cannot be zero). Therefore, $\phi_X\phi_K=\phi_Y\phi_K\Leftrightarrow\phi_X=\phi_Y$.

By the uniqueness of the characteristic function \citep[Theorem 4.3]{kallenberg2006foundations}, we have
$$\phi_X=\phi_Y\Leftrightarrow \mathbb{P}_X =\mathbb{P}_Y.$$
Using the results of the two steps, we can conclude
$$\sfdiv{\mathbb{P}_X}{\mathbb{P}_Y}=0\Leftrightarrow \mathbb{P}_{\tilde{X}}=\mathbb{P}_{\tilde{Y}}\Leftrightarrow\mathbb{P}_X =\mathbb{P}_Y.$$
\end{proof}

\section{Spread Noise Makes Distributions More Similar\label{app:dpi}}

The data processing inequality for $f$-divergences \citep{fanoinequality} states that $\fdiv{\s{p}(y)}{\s{q}(y)}\leq \fdiv{p(x)}{q(x)}$.
For completeness, we provide here an elementary proof of this result.
We consider the following joint distributions with densities
\beq
q(y,x)=p(y{\mid}x)q(x), \ocm p(y,x)=p(y{\mid}x)p(x),
\eeq
whose marginals are the spreaded distributions
\beq
\s{p}(y)=\int  p(y{\mid}x)p(x)dx, \ocm \s{q}(y)=\int p(y{\mid}x)q(x)dx.
\eeq
%
%
The divergence between the two joint distributions is
\begin{align}
\fdiv{p(y,x)}{q(y,x)} &= \int  q(y,x)f\br{\frac{p(y{\mid}x)p(x)}{p(y{\mid}x)q(x)}}dxdy
= \fdiv{p(x)}{q(x)}.
\end{align}
%

More generally, the $f$-divergence between two marginal distributions is no larger than the $f$-divergence between the joint \citep{zhang2018}. To see this, consider
%
\begin{align}
\fdiv{p(u, v)}{q(u, v)} &= \int q(u) \int q(v{\mid}u) f\br{\frac{p(u, v)}{q(u, v)}} dy du \\
&\geq  \int q(u)  f\br{ \int q(v{\mid}u) \frac{p(u, v)}{q(v{\mid}u)q(u)} dv} du \\
& = \int  q(u) f\br{\frac{p(u)}{q(u)}} du  = \fdiv{p(u)}{q(u)}.
\end{align}
%
%
%
Hence, 
\beq
\sfdiv{q(x)}{p(x)}\equiv
\fdiv{\s{p}(y)}{\s{q}(y)}\leq \fdiv{p(y,x)}{q(y,x)} = \fdiv{p(x)}{q(x)}.
\eeq
Intuitively, spreading two distributions increases their overlap, reducing the divergence.  When the distributions $\mathbb{P}$ and $\mathbb{Q}$ are absolutely continuous and their densities $p$ and $q$ have the same support, the spread $f$-divergence is always a lower bound of $f$-divergence. When the densities do not have the same support or are not well defined, then $\fdiv{\mathbb{P}}{\mathbb{Q}}$ is not well-defined.


\section{Mixture Divergence\label{app:mixture:divergence}}
We motivated the Spread Divergence between distribution $\mathbb{P}$ and $\mathbb{Q}$ by the requirement to produce a divergence that satisfying $\sdiv{\mathbb{P}}{\mathbb{Q}} = 0 \Rightarrow \mathbb{P}=\mathbb{Q}$, where the original $\div{\mathbb{P}}{\mathbb{Q}}$ does not exist. We briefly discuss the case that  $\mathbb{P}$ and $\mathbb{Q}$ are absolutely continuous but their density functions $p$ and $q$ have different supports, so $f$-divergence $\fdiv{\mathbb{P}}{\mathbb{Q}}=\div{p}{q}$ is still not defined. For example, $\mathbb{P}$ and $\mathbb{Q}$ can be two uniform distributions with different supports. We mention here an alternative divergence that also can be used , namely a mixture divergence, and discuss why we focus on the Spread Divergence thereafter.  Specifically, we can define a mixture model with density $\tilde{p}(x)$ of the original distribution and a `noise' distribution with density function $n(x)$:
\beq
\tilde{p}(x) = \alpha p(x) + (1-\alpha)n(x)
\eeq
for $0<\alpha<1$.  Provided $n(x)$ is non-zero, then $\tilde{p}(x)$ has support everywhere. Similarly, we can define
\beq
\tilde{q}(x) = \alpha q(x) + (1-\alpha)n(x).
\eeq
As with the Spread Divergence formulation presented previously, this will usually enable us to define a divergence $\div{\tilde{p}}{\tilde{q}}$ when $\supp{p}\neq\supp{q}$. Furthermore, provided the divergence between $\tilde{p}$ and $\tilde{q}$ is zero, then the two distributions $\tilde{p}$ and $\tilde{q}$ match, as do the original distributions $p$ and $q$ since
\beq
\tilde{p}(x) = \tilde{q}(x) \Leftrightarrow  \alpha p(x) + (1-\alpha)n(x) =\alpha q(x) + (1-\alpha)n(x)  \Leftrightarrow p(x)=q(x).
\eeq
Therefore, creating a mixture model in this way also allows us to define a divergence between absolutely continuous distributions that otherwise would not have an appropriate divergence\footnote{This approach is equivalent to the `anti-freeze' method used in \cite{furmston-barber-09}, which was used to enable EM style training in deterministic transition Markov Decision Processes of discrete states - see also \cite{Barber:2012:BRM:2207809}.}. However, in contrast to the Spread Divergence formulation, we cannot use this approach for distributions that are not absolutely continuous, which for many applications of interest cannot be achieved. As a simple example, consider generalised densities $p(x)=\delta\br{x-\mu_p}$,  $q(x)=\delta\br{x-\mu_q}$ with
\beq
\tilde{p}(x) = \alpha \delta\br{x-\mu_p} + (1-\alpha)n(x), \ocm \tilde{q}(x) = \alpha \delta\br{x-\mu_q} + (1-\alpha)n(x). 
\eeq
In this case, the divergence $\div{\tilde{p}(x)}{\tilde{q}(x)}$ is not defined since neither $\tilde{p}(x)$ nor $\tilde{q}(x)$ is a valid probability density. A similar issue arises in training implicit generative models in which a  value cannot be feasibly computed for $\tilde{p}$ or $\tilde{q}$;  see \secref{sec:delta:vae}. Hence, for implicit models in, we cannot feasibly assign a value to this mixture divergence. As such it appears to have only limited value in training continuous variable models.

One can combine the spread and the mixture approaches to produce a more general affine divergence
\beq
\tilde{p}(y) = \alpha \int p(y|x)p(x)dx + (1-\alpha)n(y),
\label{eq:affine}
\eeq
for spread $p(y|x)$ and (generalised) density $p(x)$. It follows for this case that $\div{\tilde{p}}{\tilde{q}}=0\Leftrightarrow \mathbb{P}=\mathbb{Q}$; however, the benefit of the mixture noise over the spread noise is not clear.
Our central interest in this work is to train implicit models and, as such, we focus interest only on the first `spread' term $\int_x p(y|x)p(x)$ in \eqref{eq:affine} and leave the study of the potential additional benefits of including a mixture component $n(y)$ for future work.

\section{Statistical Properties of Maximum Likelihood Estimator\label{app:statistical_property}}
\subsection{Existence of Spread MLE\label{app:mle:not:defined}}

In some situations there may not exist a Maximum Likelihood Estimator (MLE) for $p(x|\theta)$, but there can exist a MLE for the spread model $p(y|\theta)=\int p(y|x)p(x|\theta)dx$. For example, 
suppose that $X\sim \mathcal{N}(\mu,\sigma^2)$ ($\mu,0<\sigma^2<\infty$). So $\theta=(\mu,\sigma^2)\in \mathbb{R}\times \mathbb{R}^{+}$. Assume we only have one data point $x$. Then the log-likelihood function is $L(x;\theta)\propto -\log\sigma-\frac{1}{2\sigma^2}(x-\mu)^2$. Maximising with respect to $\mu$, we have $\mu=x$ and the log-likelihood becomes unbounded as $\sigma^2\rightarrow 0$. In this sense, the MLE for $(\mu,\sigma^2)$ does not exist, see \cite{casella2002statistical} for more discussions.

In contrast, we can check whether the MLE for $p(y|\theta)$ exists. We  assume Gaussian spread noise with fixed variance $\sigma^2_f$. Since we only have one data point $x$, the spread data distribution becomes $p(y|x)=\mathcal{N}(y|x,\sigma^2_{f})$, and the model is $p(y|\theta)=\mathcal{N}(y|\mu,\sigma^2+\sigma_f^2)$. We can sample $N$ points from the spread model, so the spread log likelihood function is (neglecting constants) $L(y_1,\ldots,y_N;\theta)= -\frac{N}{2}\log(\sigma^2+\sigma_f^2)-\frac{1}{2(\sigma^2+\sigma_f^2)}\sum_{i=1}^N(y_i-\mu)^2$. The MLE solution for $\mu$ is $\mu=\frac{1}{N}\sum_{i=1}^N y_i$; the MLE solution for $\sigma^2$ is $\sigma^2=\frac{1}{N}\sum_i (y_i-\mu)^2-\sigma_f^2$, which has bounded spread likelihood value. Note that in the limit of a large number of spread samples $N\rightarrow\infty$ , the MLE  $\sigma^2=\frac{1}{N}(y_i-\mu)^2\rightarrow\sigma_f^2$ tends to 0. Throughout, however, the (scaled by $N$) log likelihood remains bounded.

\subsection{Consistency}

  Consistency of an estimator is an important property that guarantees the validity of the resulting estimate at convergence as the number of data points tends to infinity. In what follows, we outline the sufficient conditions for a consistent MLE estimator, before addressing the question of whether using spread MLE is also consistent and under what conditions.

\subsubsection{Consistency for MLE}
Sufficient conditions for the MLE being consistent and converging to the \textit{global} maximum are given in \cite{wald1949note}. However, they are usually difficult to check even for some standard distributions. The sufficient conditions for MLE being consistent and converging to a \textit{local} maxima are given in \cite{cramer1999mathematical} and are more straight forward to check:

\begin{itemize}
\item[C1.] (Identifiable): $p(x|\theta_1)=p(x|\theta_2) \rightarrow \theta_1=\theta_2$.
\item[C2.] The parameter space $\Theta$ is an open interval $(\ubar{\theta},\bar{\theta})$, $\Theta: -\infty\leq\ubar{\theta}<\theta<\bar{\theta}\leq\infty$.
\item[C3.] $p(x|\theta)$ is continuous in $\theta$ and differentiable with respect to $\theta$ for all x.
\item[C4.] The set $A=\{x: p_\theta(x)>0\}$ is independent of $\theta$.
\end{itemize}

Let $X_1,X_2,\ldots$ be $i.i.d$ with density $p(x|\theta_0)$ ($\theta\in \Theta$) satisfying conditions C1--C4, then there exists a sequence $\hat{\theta}_n=\hat{\theta}_n(X_1,...,X_n)$ of local maxima of the likelihood function $L(\theta_0)=\prod_{i=1}^n p(x_i|\theta_0)$ which is consistent:
\begin{align*}
   \hat{\theta}\xrightarrow{p}\theta_0\quad \text{for all } \theta\in\Theta.
\end{align*}
The proof can be found in \cite{lehmann2004elements} or \cite{cramer1999mathematical}.

\subsubsection{Consistency of spread MLE}
We provide the necessary conditions for Spread MLE being consistent.
\begin{itemize}
\item[C1.] (Identifiable): $p(x|\theta)$ is identifiable. From \secref{sec:stat} it follows immediately that $p(y|\theta_1)=p(y|\theta_2) \rightarrow p(x|\theta_1)=p(x|\theta_2)\rightarrow 
\theta_1=\theta_2$, where the final implication follows from the assumption that  $p(x|\theta)$ is identifiable. Hence if $p(x|\theta)$ is identifiable, so is $p(y|\theta)$.
\item[C2.] The parameter space $\Theta$ is an open interval $(\ubar{\theta},\bar{\theta})$, $\Theta: -\infty\leq\ubar{\theta}<\theta<\bar{\theta}\leq\infty$. This condition is unchanged for $p(y|\theta)$.
\item[C3.] On $p(y|\theta)$, we require the same condition on $p(x|\theta)$ as in MLE; $p(y|\theta)$ is continuous in $\theta$ and differentiable with respect to $\theta$ for all $y$. 
\item[C4.] For spread noise $p(y|x)$ who has full support on $\mathbb{R}^d$ (for example Gaussian noise), $p(y|\theta)$ is greater than zero everywhere and hence the original condition C4 is automatically guaranteed.
\end{itemize}
The conditions that guarantee  consistency for spread MLE are weaker for the spread model $p(y|\theta)$ than for the standard model $p(x|\theta)$, since C4 is automatically satisfied. \citep{ferguson1982inconsistent} gives an example for which MLE exists but is not consistent by violating condition C4, whereas spread MLE can be  used to obtain a consistent estimator.

\subsection{Asymptotic Efficiency}

A key desirable property of any estimator is that it is efficient. The Cramer-Rao bound places a lower bound on the variance of any unbiased estimator and an efficient estimator must reach this minimal value in the limit of a large amount of data. Under certain conditions (see below) the Maximum Likelihood Estimator attains this minimal variance value meaning that there is no better estimator possible (in the limit of a large amount of data). This is one of the reasons that the maximum likelihood is a cherished criterion. 
\subsubsection{Asymptotic Efficiency for MLE} Building upon conditions C1-C4, additional conditions on $p(x|\theta)$ are required to show MLE is asymptotical efficient: 
\begin{itemize}
    \item[C5.] For all $x$ in its support, the density $p_\theta(x)$ is three times differentiable with respect to $\theta$ and the third derivative is continuous.
    \item[C6.] The derivatives of the integral $\int p_\theta(x)dx$ respect to $\theta$ can be obtained by differentiating under the integral sign, that is: $\nabla_\theta\int p_\theta(x)dx=\int \partial_\theta p_\theta(x)dx$.
    \item[C7.] There exists a positive number $c(\theta_0)$  and a function $M_{\theta_0}(x)$ such that 
    \begin{align*}
       \left|\frac{\partial^3}{\partial\theta^3}\log p_\theta(x)\right|\leq M_{\theta_0}(x) \quad \text{for all } x\in A, \left|\theta-\theta_0\right|<c(\theta_0),
    \end{align*}
    where $A$ is the support set of $x$ and $\mathbb{E}_{\theta_0}\left[M_{\theta_0}(x)\right]<\infty$.
\end{itemize}
Let $X_1,...,X_n$ be $i.i.d$ with density $p_\theta(x)$ and satisfy conditions C1-C7,
then any consistent sequence $\hat{\theta}=\hat{\theta}_n\left(X_1,...,X_n\right)$ of roots of the likelihood equation satisfies
\begin{align}
    \sqrt{n}(\hat{\theta}-\theta_0)\xrightarrow{d}\mathcal{N}\left(0,F(\theta_0)^{-1}\right),
\end{align}

where $F^{-1}(\theta_0)$ is the inverse of Fisher information matrix (also called Cramér-Rao Lower Bound, which is a lower bound on variance of any unbiased estimators ). The conditions and proof can be found in \citep{lehmann2004elements}.

\subsubsection{Asymptotic Efficiency for MLE}
As with MLE above, we require further conditions on $p(y|\theta)$ for ensuring spread MLE is asymptotically efficient:
\begin{itemize}
    \item[C5.] On $p(y|\theta)$, we require the same condition as applied to $p(x|\theta)$ in the MLE case; for all $y$ in its support, the density $p_\theta(y)$ is three times differentiable with respect to $\theta$ and the third derivative is continuous.
    
    \item[C6.] For spread noise $p(y|x)$, which has full support on $\mathbb{R}^d$ (for example Gaussian noise), the support of $y$ is independent of $\theta$. Leibniz's rule\footnote{Leibniz's rule tells us: $\frac{d}{d\theta}\int_{a(\theta)}^{b(\theta )}p(x,\theta)dx=\int_{a(\theta)}^{b(\theta)}\partial_\theta p(x,\theta)dx+p(b(\theta),\theta)\frac{d}{d\theta}b(\theta)-p(a(\theta),\theta)\frac{d}{d\theta}a(\theta)$, so if $a(\theta)$ and $b(\theta)$ are independent of $\theta$, then $\frac{d}{d\theta}\int_{a}^{b }p(x,\theta)dx=\int_{a}^{b}\partial_\theta p(x,\theta)dx$.} allows us to differentiate under the integral: $\nabla_\theta\int p_\theta(y)dy=\int \partial_\theta p_\theta(y)dy$, so this condition is automatically satisfied.
    \item[C7.] On $p(y|\theta)$, we require the same condition as applied to $p(x|\theta)$ in the MLE case; There exist positive number $c(\theta_0)$  and a function $M_{\theta_0}(y)$ such that 
    \begin{align*}
        \left|\frac{\partial^3}{\partial\theta^3}\log p_\theta(y)\right|\leq M_{\theta_0}(y) \quad \text{for all } y\in A, \left|\theta-\theta_0\right|<c(\theta_0),
    \end{align*}
    where $A$ is the support set of $y$ and $\mathbb{E}_{\theta_0}\left[M_{\theta_0}(y)\right]<\infty$.
\end{itemize}
Thus the conditions that guarantee asymptotic efficiency for the spread model $p(y|\theta)$ are weaker than for the standard model $p(x|\theta)$, since C4 and C6 are automatically satisfied.

\section{Spread Divergence for Deterministic Deep Generative Models\label{app:nn}}

Instead of minimising the likelihood, we train an implicit generative model by minimising the Spread Divergence
\beq
\min_\theta \KL{\s{p}(y)}{\s{p}_\theta(y)}.
\eeq
For Gaussian noise with fixed diagonal noise $p(y|x)=N(y|x,\sigma^2I_X)$, we can write
\beq
\s{p}(y) = \frac{1}{N}\sum_{n=1}^N \ndist{y}{x_n}{\sigma^2 I_X}.
\eeq
and 
\beq
\s{p}_\theta(y) = \int  p(y{\mid}x)p_\theta(x) dx = \int  \ndist{y}{g_\theta(z)}{\sigma^2I_X}p(z) dz = \int  p_\theta(y{\mid}z)p(z) dz.
\eeq
For the Spread Divergence with learned covariance Gaussian noise, which is discussed in \secref{sec:learn_cov}, we can write \beq 
p_\psi(y|x)=\mathcal{N}\left(y|x,\Sigma_\psi\right) , \ocm \s{p}(y) = \frac{1}{N}\sum_{n=1}^N \ndist{y}{x_n}{\Sigma_\psi}
\eeq
and Spread Divergence with a learned injective function as discussed in \secref{sec:injective}
\beq 
p_\psi(y|x)=\mathcal{N}\left(y|f_\psi(x),\sigma^2 I_X\right) , \ocm \s{p}(y) = \frac{1}{N}\sum_{n=1}^N \mathcal{N}\left(y|f_\psi(x),\sigma^2 I_X\right).
\eeq

According to our general theory,
\beq
\min_\theta \KL{\s{p}(y)}{\s{p}_\theta(y)} = 0 \hcm\Leftrightarrow\hcm p(x) = p_\theta(x).
\eeq
Here
\beq
\KL{\s{p}(y)}{\s{p}_\theta(y)}  = - \int \s{p}(y) \log \s{p}_\theta(y)dy  + \const
\eeq
Typically, the integral over $y$ will be intractable and we resort to an unbiased sampled estimate (though see below for Gaussian $q$). Neglecting constants, the KL divergence estimator is
\beq
\frac{1}{NS}\sum_{n=1}^N\sum_{s=1}^S \log \s{p}_\theta(y^n_s),
\eeq
where $y^n_s$ is a perturbed version of $x_n$.  For example $y^n_s\sim \ndist{y^n_s}{x_n}{\sigma^2I_X}$ for the fixed Gaussian noise case.
In most cases of interest, with non-linear $g$, the distribution $\s{p}_\theta(y)$ is intractable. We therefore use the variational lower bound
\begin{align}
  \log \s{p}_\theta(y) \geq \int q_\phi(z{\mid}y)\br{-\log q_\phi(z{\mid}y) + \log\br{p_\theta(y{\mid}z)p(z)}} dz.
\end{align}

Parameterising the variational distribution as a Gaussian,
\begin{align}
   q_\phi(z{\mid}y)=\ndist{z}{\mu_\phi(y)}{\Sigma_\phi(y)},
\end{align}
we can then use the reparameterization trick \citep{vae} and write
\begin{align}
    \log \s{p}_\theta(y) \geq H(\Sigma_\phi(y)) + \ave{\log p_\theta(y{\mid}z=h_\phi(y,\epsilon))+\log p(z=h_\phi(y,\epsilon))}{\ndist{\epsilon}{0}{I}},
\end{align}

where $h_\phi(y,\epsilon)=\mu_\phi(y)+C_\phi(y)\epsilon$ and $H(\Sigma_\phi(y))$ is the entropy of a Gaussian with covariance $\Sigma_\phi(y)$, where $C_\phi(y)$ is the Cholesky decomposition of $\Sigma_\phi(y)$. For fixed covariance Gaussian spread noise in $D$ dimensions, this is (ignoring the constant)

\begin{align}
  \log\s{p}_\theta(y) \geq H(\Sigma_\phi(y))
   +\ave{- \frac{1}{\br{2\sigma^2}^{D/2}}\br{y- g_\theta\br{h_\phi(y,\epsilon)}}^2 + \log p(z=h_\phi(y,\epsilon))}{\ndist{\epsilon}{0}{I}}.
 \end{align}

We can integrate the above equation over $y$ to give the bound (ignoring the constant)

\begin{align}
\int \ndist{y}{x}{\sigma^2I_X}  \log \s{p}_\theta(y) &\geq  \ave{H(\Sigma_\phi(y))+\ave{\log p(z=h_\phi(y,\epsilon))}{\ndist{\epsilon}{0}{I}}}{\ndist{y}{x}{\sigma^2I_X}}\nonumber\\
&\quad - \frac{1}{\br{2\sigma^2}^{D/2}}\ave{\ave{\br{y- g_\theta\br{h_\phi(y,\epsilon)}}^2}{\ndist{y}{x}{\sigma^2I_X}}}{\ndist{\epsilon}{0}{I}},
\end{align}
where
\begin{align}
\mathbb{E}_{\ndist{y}{x}{\sigma^2I_X}}\big[ (y- g_\theta\br{h_\phi(y,\epsilon)})^2 \big]
&=\sigma^2-2\ave{\epsilon_x g_\theta(h_\phi(x+\sigma\epsilon_x,\epsilon))}{\ndist{\epsilon_x}{0}{I_X}}\nonumber
\\&\quad +\ave{ \br{x - g_\theta(h_\phi(x+\sigma\epsilon_x,\epsilon))}^2}{\ndist{\epsilon_x}{0}{I_X}}.
\end{align}
We notice that the second term is zero, so the final bound for the fixed Gaussian spread KL divergence is (ignoring the constant)


\begin{align}
\int &\ndist{y}{x}{\sigma^2I_X}\log \s{p}_\theta(y) \geq\ave{H(\Sigma_\phi(y))+\ave{ \log p(z=h_\phi(y,\epsilon))}{\ndist{\epsilon}{0}{I}}}{\ndist{y}{x}{\sigma^2I_X}}\nonumber\\
&- \frac{1}{\br{2\sigma^2}^{D/2}}\ave{\ave{ \br{x - g_\theta(h_\phi(x+\sigma\epsilon_x,\epsilon))}^2}{\ndist{\epsilon}{0}{I}}}{\ndist{\epsilon_x}{0}{I_X}}.
\label{eq:fix_spread_bound}
\end{align}

By analogy, for spread KL divergence with learned variance, the bound is (ignoring the constant)

\begin{align}
\int &\ndist{y}{x}{\Sigma_\psi}\log \s{p}_\theta(y) \geq \ave{H(\Sigma_\phi(y))+\ave{\log p(z=h_\phi(y,\epsilon))}{\ndist{\epsilon}{0}{I}}}{\ndist{y}{x}{\Sigma_\psi}}\nonumber\\
&- \ave{\ave{ \br{x - g_\theta(h_\phi(x+S_\psi\epsilon_x,\epsilon))}^T\Sigma_\psi^{-1}\br{x - g_\theta(h_\phi(x+S_\psi\epsilon_x,\epsilon))}}{\ndist{\epsilon}{0}{I}}}{\ndist{\epsilon_x}{0}{\Sigma_\psi}},
\end{align}
where $S_\psi$ is the cholesky decomposition of $\Sigma_\psi$.


For spread KL divergence with a learned injective function, the bound is (ignoring the constant)

\resizebox{.95\hsize}{!}
{
\begin{minipage}{\linewidth}
\begin{align}
\int &\ndist{y}{f_\psi(x)}{\sigma^2I_X}\log \s{p}_\theta(y) \geq \ave{H(\Sigma_\phi(y))+\ave{\log p(z=h_\phi(y,\epsilon))}{\ndist{\epsilon}{0}{I}}}{\ndist{y}{x}{\sigma^2I_X}}\nonumber\\
&- \frac{1}{\br{2\sigma^2}^{D/2}}\ave{\ave{ \br{f_\psi(x) - f_\psi( g_\theta(h_\phi(x+\sigma\epsilon_x,\epsilon)))}^2}{\ndist{\epsilon}{0}{I}}}{\ndist{\epsilon_x}{0}{I_X}}.
\end{align}
 \end{minipage}
}

The overall procedure is therefore a straightforward modification of the standard VAE method \citep{vae} with an additional sub-routine for learning the spread online to maximize the divergence:
\begin{enumerate}
	\item Choose a noise distribution $p(y|x)$.
	\item Choose a tractable family for the variational distribution, for example $q_\phi(z{\mid}y)=\ndist{z}{\mu_\phi(y)}{\Sigma_\phi(y)}$, and initialise $\phi$.
	\item Sample a $y_n$ for each data point (if we're using $S=1$ samples).
	\item If learning the spread noise:
	\begin{enumerate}
	    \item Draw samples $\epsilon$  to estimate $-\log \s{p}_\theta(y_n)$ according to the corresponding bound. 
	    \item Do a gradient ascent step in $\psi$.
	\end{enumerate}
	\item Draw samples $\epsilon$ to estimate $\log \s{p}_\theta(y_n)$ according to the corresponding bound. 
	\item Do a gradient ascent step in $(\theta,\phi)$.
	\item Go to 3 and repeat until convergence.
\end{enumerate}

\section{MNIST Experiment\label{app:mnist}}

We first scaled the MNIST data to lie in $[0, 1]$. We use Laplace spread noise with $\sigma=0.3$ and Gaussian spread noise with $\sigma=0.3$ for the $\delta$-VAE case. Both the encoder and the decoder networks contain 3 feed-forward layers, each layer has 400 units and use ReLu activation functions. The latent dimension is $Z=64$. The variational inference network $q_\phi(z{\mid}y)=\mathcal{N}(z |\mu_\phi(y), \sigma^2_\phi I_Z)$ has a similar structure for the mean network $\mu_\phi(y)$. For fixed spread $\delta$-VAE , learning was done using the Adam \citep{kingma2014adam} optimizer with learning rate $5e^{-4}$ for 200 epochs. For $\delta$-VAE with learned spread (learned covariance), we interleave 2 covariance training epochs with 10 model training epochs (using the Adam optimizer with learning rate $5e^{-5}$). 



\section{CelebA Experiment\label{app:celeba}}
We pre-processed CelebA images by first taking 140x140 centre crops and then resizing to 64x64.  Pixel values were then rescaled to lie in $[0,1]$. For the learned spread we use Gaussian noise with a learned injective function ResNet $f_\psi(\cdot)=I(\cdot)+g_\psi(\cdot)$, where $g_\psi(\cdot)$ is a one layer convolutional neural net with kernel size $3\times 3$, with stride length 1. We use spectral normalization \citep{miyato2018spectral} to satisfy the Lipschitz constraint. That is, we replace the  weight matrix $w$ of the convolution kernel by $w_{SN}(w) := c\times w/\sigma(w)$, where $\sigma(w)$ is the spectral norm of $w$ and $c\in (0,1)$. This guarantees that $f_\psi$ is invertible - see \cite{behrmann2018invertible}.

The encoder and decoder are 4-layer convolutional neural networks with batch norm \citep{ioffe2015batch}. Both networks use a fully convolutional architecture with 5x5 convolutional filters with stride length 2 in both vertical and horizontal directions, except the last deconvolution layer where we use stride length 1. $\text{Conv}_k$ represents a convolution with $k$ filters and $\text{DeConv}_k$ represents a deconvolution with k filters, BN for the batch normalization \citep{ioffe2015batch}, Relu for the rectified linear units, and $\text{FC}_k$ for the fully connected layer mapping to $\mathbb{R}^k$.
\begin{align*}
x\in \mathbb{R}^{64\times64\times3}&\rightarrow \text{injective}f(\cdot)\in \mathbf{R}^{64\times64\times3}\\
&\rightarrow \text{Conv}_{128}\rightarrow\text{BN}\rightarrow\text{Relu}\\
&\rightarrow \text{Conv}_{256}\rightarrow\text{BN}\rightarrow\text{Relu}\\
&\rightarrow \text{Conv}_{512}\rightarrow\text{BN}\rightarrow\text{Relu}\\
&\rightarrow \text{Conv}_{1024}\rightarrow\text{BN}\rightarrow\text{Relu}\rightarrow\text{FC}_{100}
\end{align*}

\begin{align*}
z\in \mathbb{R}^{100}&\rightarrow \text{FC}_{10\times 10\times 1024}\\ &\rightarrow\text{DeConv}_{512}\rightarrow\text{BN}\rightarrow\text{Relu}\\
&\rightarrow \text{DeConv}_{256}\rightarrow\text{BN}\rightarrow\text{Relu}\\
&\rightarrow \text{DeConv}_{128}\rightarrow\text{BN}\rightarrow\text{Relu}\rightarrow\text{DeConv}_{3}\rightarrow sigmoid(\cdot)\\
&\rightarrow \text{injective}f(\cdot)\in \mathbb{R}^{64\times64\times3}
\end{align*}

We use batch size $100$ and latent dimension $\sz{Z}=100$ in all CelabA experiments.
For the $\delta$-VAE with fixed spread, we use the fixed Gaussian noise with 0 mean and $(0.5)^2I$ covariance. We train the model for 500 epochs using Adam optimizer with learning rate $1e^{-4}$. We decay the learning rate with scaling factor $0.9$ every $100000$ iterations. 

For the $\delta$-VAE with learned spread we first train a $\delta$-VAE with fixed $f(x)=x$ and fixed Gaussian noise with 0 mean and $(0.5)^2I$ diagonal covariance  for 300 epochs. We decay the learning with scaling factor $0.9$ every $100000$ iterations. We start iterative training by doing one step inner maximisation over the Spread Divergence parameters $\psi$ using Adam optimizer with learning rate $1e^{-5}$ and one step minimization over the model parameter's $(\theta,\phi)$ using Adam optimizer for additional 200 epochs. We can share the first 300 epochs between the two models. When we sample form two models, we first sample from a $100$ dimensional standard Gaussian distribution $z\sim \mathcal{N}(0,I)$ and use the same latent code $z$ to get samples from both $\delta$-VAE with fixed and learned spread, so we can easily compare the sample quality between two models. 


\begin{figure}[h]
    \centering
    \subfloat[Laplace with fixed covariance]{
        \includegraphics[width=0.4\linewidth,valign=c]{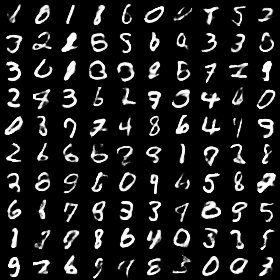}
        }
    \subfloat[Gaussian with fixed covariance]{
        \includegraphics[width=0.4\linewidth,valign=c]{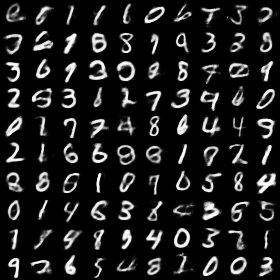}
        }\\
    \subfloat[Gaussian with learned covariance]{
        \includegraphics[width=0.4\linewidth,valign=c]{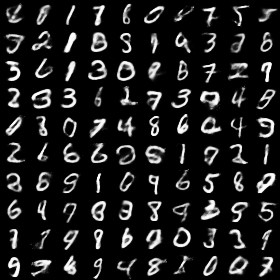}
        }
    \caption{Samples from an implicit generative model trained using $\delta$-VAE with (a) Laplace noise with fixed covariance, (a) Gaussian noise with fixed covariance and (c) Gaussian noise with learned covariance.
    \label{fig:app:mnist}}
\end{figure}
\begin{figure}[t]
    \centering
    \subfloat[Fixed spread noise]{
        \includegraphics[width=0.6\linewidth,valign=c]{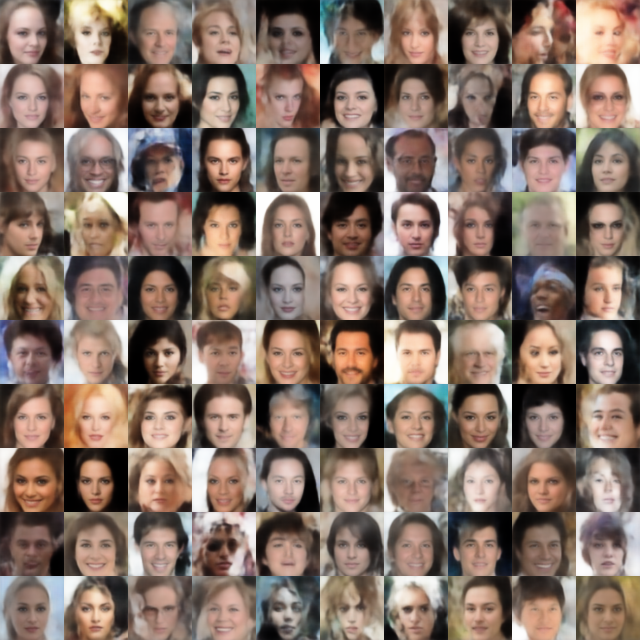}
        }\\
    \subfloat[Learned spread noise]{
        \includegraphics[width=0.6\linewidth,valign=c]{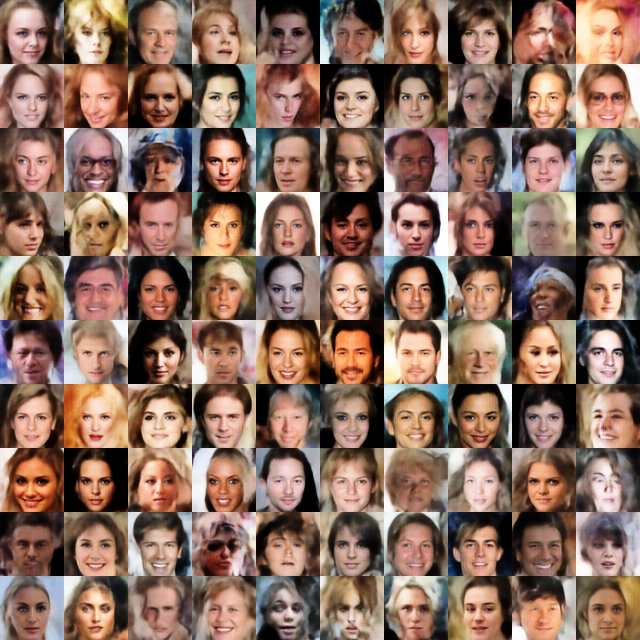}
        }
    \caption[]{Samples from an implicit generative model trained using $\delta$-VAE with (a) fixed and (b) learned spread with injective mean transformation.\label{fig:more:samples:celeb}}
\end{figure}

\end{document}